\documentclass[11pt]{article}
\usepackage[T1]{fontenc}
\usepackage[utf8]{inputenc}
\usepackage{lmodern}
\usepackage[a4paper,margin=23mm]{geometry}
\usepackage{amsmath,amssymb,amsthm,booktabs,longtable,array,enumitem,algorithm,algorithmic}

\usepackage{graphicx,needspace}
\graphicspath{{icml2027/}}
\usepackage{hyperref}
\hypersetup{hidelinks,
 pdftitle={Anytime-Valid Model Evaluation with Selective Queries: Top-K Identification and Target Qualification},pdfauthor={Qi Kuang and Yin Xia}}
\setlist{nosep,leftmargin=1.6em,topsep=3pt}
\newcommand{\E}{\mathbb E}
\newcommand{\Prb}{\mathbb P}
\newcommand{\R}{\mathbb R}

\newcommand{\Cost}{\operatorname{Cost}}
\newcolumntype{L}[1]{>{\raggedright\arraybackslash}p{#1}}
\newtheorem{theorem}{Theorem}
\newtheorem{proposition}{Proposition}
\newtheorem{lemma}[theorem]{Lemma}

\begin{document}

\begin{center}
{\Large\bfseries Anytime-Valid Model Evaluation with Selective Queries:\\[2pt]
Top-$K$ Identification and Target Qualification}\\[6pt]
Qi Kuang and Yin Xia\\[2pt]
{\small Department of Statistics and Data Science, Fudan University, Shanghai, China}
\end{center}

\begin{abstract}
Many model evaluations only need to identify the Top-$K$ models or decide whether a target model belongs to that set, rather than recover a complete ranking. Running every model on every test question is costly. We introduce PACE, a sequential evaluation framework covering Top-$K$ identification and target qualification under both full observation and selective querying. PACE tests each candidate hypothesis by projecting past mean estimates onto a convex cone to form a difference between two weighted averages of fresh scores on the same question. This payoff cancels any additive question effect shared by the models and updates an e-process via Follow-the-Regularized-Leader betting to eliminate false candidates. Under selective querying, PACE uses the remaining candidates and current estimates to choose which models answer each new question, skipping models when they are not needed and querying them again if uncertainty returns. Allowing arbitrary dependence between scores on the same question, we prove finite-sample anytime-valid error control and, whenever the relevant mean gap is positive, almost sure termination with finite expected query cost for all four methods. We also quantify the savings from selective querying and from testing a target directly. On six Open LLM Leaderboard benchmarks with pools of 15 models, full-observation PACE uses $41.3\%$ to $46.5\%$ fewer questions on average than RankCS for Top-5 identification and $34.6\%$ to $53.6\%$ fewer for target qualification. Relative to RankCS with model retirement, selective PACE reduces mean total model queries by $35.5\%$ to $38.4\%$ for Top-5 identification and by $45.6\%$ to $68.4\%$ for target qualification.
\end{abstract}
\section{Introduction}

Evaluating $M$ large language models or other machine learning systems can require many test questions and substantial compute or API cost~\cite{polo2024tinybenchmarks,arviv2026stop}. Often, the decision does not require a complete ranking. Selecting $K$ models for traffic routing or ensembling requires knowing which models belong to the Top-$K$, without ordering models within or outside that set. We call this task \emph{Top-$K$ identification}. When evaluating a new checkpoint $J$, the question may be simpler: does $J$ belong to the Top-$K$? We call this \emph{target qualification}, with answer IN or OUT.

Sequential evaluation processes questions one at a time and stops when there is enough evidence to answer. A test designed for a fixed sample size does not generally retain its error guarantee when the stopping time is chosen from the data~\cite{armitage1969repeated,khosravi2026rank}. We therefore require \emph{anytime validity}: the error guarantee must hold whenever the evaluator decides to stop~\cite{ville1939etude,howard2021time,ramdas2023game}.

RankCS~\cite{khosravi2026rank} provides these error guarantees for sequential ranking. With its retirement rule, RankCS stops querying a model once there is enough evidence to decide whether that model belongs to the Top-$K$. Until then, the model must answer every question, even when only a few close competitors are still difficult to distinguish. RankCS also averages evidence across pairs of models. If only a few pairs show that a proposed Top-$K$ set is wrong, averaging their evidence with weaker evidence from other pairs can make that set harder to rule out.

Adaptive bandit methods such as LUCB++ and Track-and-Stop~\cite{garivier2016optimal,simchowitz2017simulator} save queries by focusing on models that are still difficult to distinguish. However, they use separate model queries or estimates of each model's mean, without directly comparing scores on the same question. Such comparisons can be useful: a hard question may lower both models' scores, while an easy one may raise both. Taking the score difference on each question can remove this shared effect and make the comparison more precise.

Our goal is to combine these advantages: choose which models answer each question, compare their scores on that question, and retain finite-sample anytime-valid error control. To do so, we introduce PACE for Top-$K$ identification and target qualification. PACE represents each candidate hypothesis by a convex cone and projects past mean estimates onto it to weight score differences on the next question. This focuses evidence on violated comparisons and cancels shared question effects. Under selective querying, the remaining candidates determine which models answer each new question, skipping resolved models and querying them again if uncertainty returns. Our contributions are as follows:
\begin{itemize}[leftmargin=1.5em,itemsep=2pt]
  \item We develop PACE for Top-$K$ identification and target qualification, under either full observation or selective querying. Cone projection determines the score comparison weights, and Follow-the-Regularized-Leader determines the betting coefficients. In each case, the finite-sample probability of an incorrect answer is at most $\alpha$ at any data-dependent stopping time, allowing arbitrary score dependence on the same question. A positive gap at the relevant cutoff ensures almost sure termination and finite expected query cost.
  \item We quantify the query savings from selective querying and direct target qualification. When a single boundary pair approaches a tie, selective querying requires the same order of questions as full observation, while reducing expected queries per question from $M$ to $2$ and reducing total expected queries to at most a $2/(M(1-2\alpha))$ asymptotic fraction of full observation. Moreover, when the target stays clearly above or below its cutoff, its expected query cost remains bounded even as the Top-$K$ boundary gap $\Delta_K$ vanishes, whereas any valid Top-$K$ identification method requires $\Omega(1/\Delta_K)$ expected queries.
  \item On six Open LLM Leaderboard benchmarks, we evaluate pools of 15 models on Top-5 identification and target qualification at ranks $1$, $3$, $8$, and $15$. Under full observation, PACE uses $41.3\%$ to $46.5\%$ fewer questions on average than RankCS for Top-5 identification and $34.6\%$ to $53.6\%$ fewer for target qualification. Relative to RankCS with retirement, selective PACE uses $35.5\%$ to $38.4\%$ fewer model queries for Top-5 identification and $45.6\%$ to $68.4\%$ fewer for target qualification. For Top-5 identification, the bandit racing baselines BRR and VRR~\cite{gharat2025representative} finish all 100 streams within budget only on GSM8K, where selective PACE uses $91.7\%$ and $89.0\%$ fewer queries, respectively.
\end{itemize}

\subsection{Related Work}

Fixed-sample rank inference methods use simultaneous pairwise comparisons to bound each model's possible ranks~\cite{al2022simultaneous,mogstad2024inference,neuhof2024confident,neuhof2026quantifying}. Other studies reduce evaluation cost through paired comparisons~\cite{miller2024adding} or small subsets of questions~\cite{polo2024tinybenchmarks}. These fixed-sample guarantees do not by themselves justify stopping based on the observed data. Anytime-valid confidence sequences~\cite{howard2021time,10.1093/jrsssb/qkad009} have been applied to active question selection for single-model evaluation~\cite{hsu2026efficient,zhou2026celeus}, whereas Arviv et al.~\cite{arviv2026stop} use group-sequential pairwise testing without simultaneous correction. For multi-model comparisons, Arnold et al.~\cite{arnold2026sequential} build variable-size sequential model confidence sets, RankCS~\cite{khosravi2026rank} applies closed testing to pairwise e-processes and allows models to retire, and SERPANT~\cite{guanytime} ranks LLMs from sequential pairwise preferences. Scalar confidence sequences~\cite{10.1093/jrsssb/qkad009} can also be combined across pairs via multiple-testing corrections. By contrast, PACE selects which models answer each shared question and tests each composite hypothesis with a single e-process.

Choosing which models to query is closely related to pure exploration in multi-armed bandits, which studies fixed-confidence arm and subset selection~\cite{even2006action,kaufmann2016complexity}, including LUCB~\cite{kalyanakrishnan2012pac}, KL-LUCB~\cite{kaufmann2013information}, LUCB++~\cite{simchowitz2017simulator}, Track-and-Stop~\cite{garivier2016optimal}, and the BRR/VRR racing procedures~\cite{gharat2025representative}. Saad et al.~\cite{saad2023covariance} jointly sample subsets of arms and use empirical variance bounds on reward contrasts to identify the best arm. PACE also uses jointly observed scores and builds an e-process for each candidate order hypothesis using a contrast obtained by cone projection. These tests support both Top-$K$ identification and direct target qualification.

To accumulate valid sequential evidence, PACE builds on test supermartingales and Ville's inequality~\cite{ville1939etude,ramdas2023game}, closed testing and e-value combination~\cite{marcus1976closed,shaffer1986modified,vovk2021values,hartog2025family}, and online convex optimization and coin-betting methods~\cite{mcmahan2011follow,shalev2025online,hazan2016introduction,orabona2016coin}.
Zhao~\cite{zhao2026approachability} connects Blackwell approachability to betting tests for convex mean hypotheses via support-function residuals, with applications including two-sample testing. By contrast, PACE tests families of order-cone hypotheses for multi-model Top-$K$ identification and target qualification, combining cone-projection contrasts with FTRL betting. Under selective querying, PACE uses the remaining hypotheses to choose which models answer each new question.

\section{Problem Formulation}\label{sec:formulation}

\subsection{Data and Queries}

One query evaluates one model on one test question. Let $X_{t,i}\in[0,1]$ be model $i$'s score on question $t$. The full score vectors
\[
 X_t=(X_{t,1},\ldots,X_{t,M})
\]
are i.i.d.\ draws from an unknown distribution $P$. Scores on the same question may be arbitrarily dependent: a difficult question may lower several models' scores together. Write
\[
 \mu_i=\E_P X_{t,i},\qquad \mu=(\mu_1,\ldots,\mu_M)
\]
for the population mean scores. A larger mean indicates better performance.

Before seeing any score on question $t$, we choose which models to query. Let $Q_t\subseteq[M]=\{1,\ldots,M\}$ be this set; only scores $X_{t,i}$ with $i\in Q_t$ are observed. The total query cost through question $t$ is
\[
 \Cost_t=\sum_{s=1}^{t}|Q_s|.
\]
Let $\mathcal F_{t-1}$ denote the information generated by the query choices, observed scores, and randomization used through question $t-1$.

\subsection{Evaluation Tasks}

Fix $K\in\{1,\ldots,M-1\}$ and an error level $\alpha\in(0,1)$ before evaluation. The required answer depends on the chosen task:
\begin{itemize}
 \item Top-$K$ identification returns the $K$ models with the largest population means.
 \item Target qualification concerns one model $J$, fixed before evaluation. It returns IN if $J$ ranks among the top $K$, and OUT otherwise.
\end{itemize}
For either task, the procedure uses the accumulated evidence to decide when to return its answer. Let $\tau$ be this data-dependent stopping time, measured in questions. We require the probability of an incorrect answer to be at most $\alpha$, while keeping both $\tau$ and the total query cost $\Cost_\tau$ small. These counts differ because one question may be sent to several models.

Either task can use full observation, with $Q_t=[M]$, or selective querying, which chooses $Q_t$ using past observations and may query fewer models. Combining the two tasks with these two query settings gives four methods:
\begin{center}\small
\begin{tabular}{lll}\toprule
Method & Requested output & Observation\\\midrule
M1 & Top-$K$ set & Full\\
M2 & Target IN/OUT & Full\\
M3 & Top-$K$ set & Selective\\
M4 & Target IN/OUT & Selective\\\bottomrule
\end{tabular}
\end{center}
Section~\ref{sec:full-methods} first develops the tests and stopping rules under full observation (M1 and M2). Section~\ref{sec:query} then adapts these procedures to selective querying (M3 and M4).

\section{Full Observation}\label{sec:full-methods}\label{sec:e-process}

Under full observation, every model answers each new question. We must decide when these scores provide enough evidence to return the requested answer while keeping the error probability at most $\alpha$. To do so, we test candidate hypotheses and rule them out as evidence accumulates. We first develop this approach for Top-$K$ identification (M1), then adapt it to the target's IN/OUT decision (M2).

\subsection{Top-\texorpdfstring{$K$}{K} Identification}\label{sec:set-method}

\subsubsection{Candidate Hypotheses}

Because the population means are unknown, we initially treat every set of $K$ models as a possible Top-$K$ set. For a candidate $S\subseteq[M]$ with $|S|=K$, define the hypothesis
\[
 H_S:\mu\in C_S,\qquad
 C_S=\{\theta\in\R^M:\theta_i\ge\theta_j
       \text{ for all }i\in S,\ j\notin S\}.
\]
This cone places every model in $S$ at or above every model outside $S$, without ordering models within either group. We reject candidates as evidence accumulates and return when one candidate remains.

Let $S^\star$ be a set of $K$ models with the largest population means, with boundary gap
\[
 \Delta_K=\min_{i\in S^\star}\mu_i-\max_{j\notin S^\star}\mu_j\ge0.
\]
When $\Delta_K>0$, $S^\star$ is the unique Top-$K$ set. Models within either group may still have equal means.

\subsubsection{E-processes}

Simply ranking models by their running sample means and taking the top $K$ does not tell us when to stop: sampling fluctuations can change the empirical Top-$K$ set across questions, and stopping at a data-dependent time invalidates fixed-sample tests. Instead, we maintain an e-process for each candidate $S$ so that the probability of ever rejecting a true hypothesis remains at most $\alpha$.

Each candidate $S$ has a wealth account $E_{S,t}$, initialized at $E_{S,0}=1$ and updated after question $t$ by
\[
 E_{S,t}=E_{S,t-1}(1+\lambda_{S,t}z_{S,t}).
\]
Validity requires a betting payoff $z_{S,t}\in[-1,1]$ with $\E[z_{S,t}\mid\mathcal F_{t-1}]\le0$ under $H_S$ and a betting coefficient $\lambda_{S,t}\in[0,1/2]$ chosen from $\mathcal F_{t-1}$. Then $1+\lambda_{S,t}z_{S,t}\ge0$ and $\E[1+\lambda_{S,t}z_{S,t}\mid\mathcal F_{t-1}]\le1$ under $H_S$, so $(E_{S,t})_{t\ge0}$ is a nonnegative test supermartingale. By Ville's inequality~\cite{ville1939etude}, whenever $H_S$ is true,
\[
 \Prb\!\left\{\sup_{t\ge0}E_{S,t}\ge1/\alpha\right\}\le\alpha.
\]
Rejecting $H_S$ once $E_{S,t}\ge1/\alpha$ bounds the probability of ever eliminating $S^\star$ by $\alpha$. Returning an incorrect set requires eliminating $S^\star$, so the probability of an incorrect output is at most $\alpha$, without a multiplicity correction across the $\binom{M}{K}$ candidates. Below, we fix an arbitrary candidate $S$ and omit $S$ from the subscripts of $E_t$, $z_t$, and $\lambda_t$.

\subsubsection{Betting Payoffs}

For candidate $S$, the payoff is a weighted average of the new scores outside $S$ minus a weighted average of the new scores inside $S$. If $H_S$ is true, every inside mean is at least every outside mean, so any such comparison has nonpositive conditional expectation when its weights are chosen from past data. If $H_S$ is false, some outside model has a higher mean than an inside model. We use past observations to concentrate weight on these violations and obtain positive payoffs that can support rejection.

Let $\widehat\mu_{t-1}$ be the mean score vector from the first $t-1$ questions, initialized at $\widehat\mu_0=\tfrac12\mathbf1$. Find the closest vector satisfying candidate $S$'s constraints:
\[
 \Pi_{C_S}\widehat\mu_{t-1}
 =\arg\min_{\theta\in C_S}\|\widehat\mu_{t-1}-\theta\|_2^2.
\]
This Euclidean projection shows how the estimates must change to agree with the candidate. Its residual is
\[
 r=\widehat\mu_{t-1}-\Pi_{C_S}\widehat\mu_{t-1}.
\]
Projection onto $C_S$ can only raise estimates inside $S$ and lower estimates outside $S$, so $r_i\le0$ for $i\in S$ and $r_j\ge0$ for $j\notin S$.

When $r\ne0$, the betting payoff is
\begin{equation}\label{eq:projection-payoff}
 z_t
 =\frac{2r^{\top}X_t}{\|r\|_1}
 =
 \sum_{j\notin S}
 \frac{2r_j}{\|r\|_1}X_{t,j}
 -
 \sum_{i\in S}
 \frac{-2r_i}{\|r\|_1}X_{t,i}.
\end{equation}
Adding the same constant to every coordinate preserves the constraints in $C_S$. This implies $\sum_{m=1}^M r_m=0$, so the positive residuals and the absolute values of the negative residuals each sum to $\|r\|_1/2$. The factor $2/\|r\|_1$ in~\eqref{eq:projection-payoff} therefore makes each group of weights sum to one. Since both weighted average scores lie in $[0,1]$, their difference lies in $[-1,1]$.

Because the weights are fixed before question $t$ and sum to one on each side, any common additive shift in all scores on that question cancels from the payoff, and under $H_S$ the ordering of population means gives $\E[z_t\mid\mathcal F_{t-1}]\le0$. When $r=0$, the estimates already satisfy the candidate's constraints, so we set $z_t=0$.

For example, take three models $\mathrm A,\mathrm B,\mathrm C$, $K=2$, and candidate $S=\{\mathrm A,\mathrm C\}$. The candidate requires both selected models to have means at least as large as model $\mathrm B$'s. If the past means are $(0.80,0.51,0.49)$, then
\[
\begin{aligned}
 \Pi_{C_S}\widehat\mu_{t-1}&=(0.80,0.50,0.50),\\
 r&=(0,0.01,-0.01),\\
 z_t&=X_{t,\mathrm B}-X_{t,\mathrm C}.
\end{aligned}
\]
Thus the next question checks whether $\mathrm B$ again outperforms $\mathrm C$. If it does, $z_t>0$ and a positive betting coefficient increases the evidence against $S$.

\Needspace{5\baselineskip}
\subsubsection{Betting Coefficients}

We choose the betting coefficient using Follow-the-Regularized-Leader (FTRL)~\cite{mcmahan2011follow,shalev2025online,hazan2016introduction}. A fixed coefficient $b$ would have earned log wealth $\sum_{s<t}\log(1+bz_s)$ on the observed payoffs. FTRL chooses the coefficient that maximizes this past log wealth with a penalty for large bets:
\begin{equation}\label{eq:ftrl-opt}
 \lambda_t=\arg\max_{0\le b\le1/2}
 \left\{\sum_{s<t}\log(1+bz_s)+\tfrac12\log(1-b^2)\right\}.
\end{equation}
The term $\tfrac12\log(1-b^2)$ discourages large bets when few observations are available. The objective is strictly concave and gives $\lambda_1=0$. Each $\lambda_t$ uses only payoffs from earlier questions and lies in $[0,1/2]$, as required for validity.

In practice, replacing $\log(1+bz_s)$ by its degree-32 Taylor polynomial avoids storing and rescanning past payoffs for each account while preserving all theoretical guarantees; see Appendix~\ref{app:ftrl-implementation}.

\Needspace{5\baselineskip}
\subsubsection{Stopping Rule}\label{sec:set-output}

Let $\mathcal A_t$ contain the candidate sets retained after question $t$, starting with all $K$-sets. Fix an order over candidates before evaluation. After updating every account, examine the candidates in that order and delete those whose account reaches $1/\alpha$. Stop deleting as soon as one candidate remains, and return it. This prevents an empty answer set if several accounts cross the threshold on the same question. Together with $Q_t=[M]$, these rules define M1.

\subsection{Target Qualification}\label{sec:membership-method}

M2 decides whether a fixed target $J$ belongs to the Top-$K$ tier. It does not need to identify the other members. Define the target's competition rank by
\[
 \operatorname{rank}(J)=1+\sum_{i\ne J}\mathbf1\{\mu_i>\mu_J\}.
\]
The true label $\ell^\star$ is IN if $\operatorname{rank}(J)\le K$, and OUT otherwise.

\subsubsection{Candidate Hypotheses}

To certify IN, we rule out every possible group of $K$ opponents that could be ahead of $J$. For each $A\subseteq[M]\setminus\{J\}$ with $|A|=K$, define
\[
 H_A^+:\mu\in C_A^+,\qquad
 C_A^+=\{\theta\in\R^M:\theta_J\le\theta_i
                         \text{ for all }i\in A\}.
\]
If $J$ is OUT, some $K$ opponents have strictly larger means, so at least one of these hypotheses is true. Rejecting every plus hypothesis therefore certifies IN.

To certify OUT, we rule out every possible group of $M-K$ opponents that could be no better than $J$. For each $B\subseteq[M]\setminus\{J\}$ with $|B|=M-K$, define
\[
 H_B^-:\mu\in C_B^-,\qquad
 C_B^-=\{\theta\in\R^M:\theta_J\ge\theta_i
                         \text{ for all }i\in B\}.
\]
If $J$ is IN, at least one of these minus hypotheses is true. Rejecting every minus hypothesis therefore certifies OUT.

The weak inequalities make each cone closed, so both families can contain true hypotheses when opponents tie with $J$. Error control still holds at ties, though the procedure may not stop unless $J$ is separated from its cutoff.

\subsubsection{Betting Payoffs}

Fix a candidate cone $C$ in either family and compute its projection residual $r=\widehat\mu_{t-1}-\Pi_C\widehat\mu_{t-1}$ from the full-observation empirical means. As in~\eqref{eq:projection-payoff}, set $z_t=2r^\top X_t/\|r\|_1$ when $r\ne0$ and $z_t=0$ otherwise. When $r\ne0$, $C$ constrains $J$ only against opponents in $A$ or $B$, so the weights $w_i=|r_i|/|r_J|$ use only past data, are nonnegative, and sum to one. The payoff simplifies to
\[
 z_t = X_{t,J} - \sum_{i\in A} w_i X_{t,i} \quad\text{for } C_A^+,
 \qquad
 z_t = \sum_{i\in B} w_i X_{t,i} - X_{t,J} \quad\text{for } C_B^-.
\]
Under a true plus hypothesis, each opponent mean is at least $\mu_J$; under a true minus hypothesis, each is at most $\mu_J$. Thus $z_t\in[-1,1]$ and $\E[z_t\mid\mathcal F_{t-1}]\le0$ under either hypothesis.

\subsubsection{Updates and Stopping Rule}\label{sec:membership-output}

Each candidate $C$ in the two families updates its account as in Section~\ref{sec:set-method}, with its betting coefficient given by~\eqref{eq:ftrl-opt} from its own past payoffs. Let $\mathcal A_t^+$ and $\mathcal A_t^-$ index the hypotheses retained after question $t$. Fix their deletion order before evaluation, with the plus family before the minus family. After updating the accounts, examine candidates in that order and reject those reaching $1/\alpha$. Stop immediately when either family becomes empty, returning
\[
 \mathrm{IN}\quad\text{if }\mathcal A_t^+=\varnothing,
 \qquad
 \mathrm{OUT}\quad\text{if }\mathcal A_t^-=\varnothing.
\]
Stopping before any further deletion prevents both families from becoming empty on the same question. These rules, together with full observation, define M2; Algorithm~\ref{alg:pace-full} summarizes M1 and M2. As in M1, returning the wrong label requires rejecting a true hypothesis, so no multiplicity correction across candidates is needed.

The same full-observation construction can test any prespecified collection of pairwise order constraints $\mu_i\ge\mu_j$. For example, it can test a proposed complete ranking $\mu_{i_1}\ge\cdots\ge\mu_{i_M}$; see Appendix~\ref{app:order-constraints}.

\begin{algorithm}[htbp]
\caption{Full-Observation PACE: M1 and M2}
\label{alg:pace-full}
\small
\begin{algorithmic}[1]
\REQUIRE Models $[M]$, cutoff $K$, error level $\alpha$; target $J$ for M2.
\STATE Initialize $\widehat\mu_0\gets\tfrac12\mathbf1$.
\STATE \textbf{M1:} Let $\mathcal A_0$ contain all $K$-subsets $S\subseteq[M]$ in a fixed order, with cones $C_S$.
\STATE \textbf{M2:} Let $\mathcal A_0^+$ and $\mathcal A_0^-$ contain all plus cones $C_A^+$ with $|A|=K$ and minus cones $C_B^-$ with $|B|=M-K$ over opponents in $[M]\setminus\{J\}$, in a fixed order with $\mathcal A_0^+$ before $\mathcal A_0^-$.
\STATE Set $E_{C,0}\gets1$ and $\lambda_{C,1}\gets0$ for every candidate cone $C$.
\FOR{$t=1,2,\ldots$}
  \STATE For each retained cone $C$, compute $r_C\gets\widehat\mu_{t-1}-\Pi_C\widehat\mu_{t-1}$.
  \STATE Query $Q_t=[M]$ and observe $X_t=(X_{t,1},\ldots,X_{t,M})$.
  \FOR{each retained cone $C$}
    \STATE Set $z_{C,t}\gets 2r_C^\top X_t/\|r_C\|_1$ if $r_C\ne0$, and $z_{C,t}\gets0$ otherwise.
    \STATE Update $E_{C,t}\gets E_{C,t-1}(1+\lambda_{C,t}z_{C,t})$ and compute $\lambda_{C,t+1}$ from $z_{C,1},\ldots,z_{C,t}$ via~\eqref{eq:ftrl-opt}.
  \ENDFOR
  \STATE Update $\widehat\mu_t\gets\frac1t\sum_{s=1}^t X_s$.
  \STATE \textbf{M1:} Set $\mathcal A_t\gets\mathcal A_{t-1}$; delete $S\in\mathcal A_t$ in order when $E_{C_S,t}\ge1/\alpha$, stopping once $|\mathcal A_t|=1$. If $|\mathcal A_t|=1$, \textbf{return} the remaining set.
  \STATE \textbf{M2:} Set $\mathcal A_t^+\gets\mathcal A_{t-1}^+$ and $\mathcal A_t^-\gets\mathcal A_{t-1}^-$; delete $C$ in order when $E_{C,t}\ge1/\alpha$, stopping once either family is empty. \textbf{Return} $\mathrm{IN}$ if $\mathcal A_t^+=\varnothing$, or $\mathrm{OUT}$ if $\mathcal A_t^-=\varnothing$.
\ENDFOR
\end{algorithmic}
\end{algorithm}

\section{Selective Querying}\label{sec:query}

Full observation evaluates every model on every question. Once some models are clearly above or below the cutoff, the remaining decision may depend on only a few comparisons. M3 and M4 use the retained hypotheses and past scores to choose which models to evaluate next. The aim is to reduce model queries while keeping the same error guarantee.

\subsection{Top-\texorpdfstring{$K$}{K} Identification}\label{sec:discrepancy-query}

M3 uses disagreements about Top-$K$ membership to decide which models need another query. It first estimates each model's mean using only its scores observed before question $t$. Let $\widehat\mu_{t-1}$ collect these estimates, and let $\widehat S_{t-1}$ contain the $K$ models with the largest estimates, breaking ties by model index.

Before observing question $t$, independently choose a full query with probability $p_t=t^{-1/2}$, setting $Q_t=[M]$. Otherwise, query
\[
 Q_t=\bigcup_{S\in\mathcal A_{t-1}}
       (\widehat S_{t-1}\mathbin\triangle S).
\]
Thus $Q_t$ queries every model whose membership differs between $\widehat S_{t-1}$ and at least one retained candidate, and skips models on which all of them agree. Full queries continue to refresh every model's estimate; the first question is fully observed because $p_1=1$.

Once $Q_t$ is chosen, compute the comparison weights before observing scores on $Q_t$. For each retained candidate $S$, restrict $\widehat\mu_{t-1}$ and $C_S$ to $Q_t$, keeping the constraints $\theta_i\ge\theta_j$ for $i\in S\cap Q_t$ and $j\in Q_t\setminus S$. Project this restricted mean vector onto the restricted cone to obtain the residual. After the queried scores arrive, use that residual to compute the payoff in~\eqref{eq:projection-payoff}, summing only over $Q_t$ and setting the payoff to zero when the residual is zero.

Update every retained candidate's wealth account and compute its next betting coefficient using~\eqref{eq:ftrl-opt}. Also update the model mean estimates from the queried scores. Apply the deletion and stopping rule in Section~\ref{sec:set-output}. These steps define M3, summarized in Algorithm~\ref{alg:pace-m3}.

\begin{algorithm}[htbp]
\caption{Selective Top-$K$ Identification: M3}
\label{alg:pace-m3}
\small
\begin{algorithmic}[1]
\REQUIRE Models $[M]$, cutoff $K$, error level $\alpha$.
\STATE Initialize $\widehat\mu_0\gets\tfrac12\mathbf1$, and let $\mathcal A_0$ contain all $K$-subsets $S\subseteq[M]$ in a fixed order, with $E_{S,0}\gets1$ and $\lambda_{S,1}\gets0$.
\FOR{$t=1,2,\ldots$}
  \STATE With probability $p_t=t^{-1/2}$, set $Q_t\gets[M]$; otherwise let $\widehat S_{t-1}$ be the $K$ models with the largest entries in $\widehat\mu_{t-1}$, breaking ties by index, and set $Q_t\gets\bigcup_{S\in\mathcal A_{t-1}}(\widehat S_{t-1}\mathbin\triangle S)$.
  \STATE For each $S\in\mathcal A_{t-1}$, restrict $\widehat\mu_{t-1}$ and $C_S$ to $Q_t$ and project to obtain residual $r_S$.
  \STATE Query $Q_t$ and observe $X_{t,Q_t}=(X_{t,i})_{i\in Q_t}$.
  \FOR{each $S\in\mathcal A_{t-1}$}
    \STATE Set $z_{S,t}\gets 2r_S^\top X_{t,Q_t}/\|r_S\|_1$ if $r_S\ne0$, and $z_{S,t}\gets0$ otherwise.
    \STATE Update $E_{S,t}\gets E_{S,t-1}(1+\lambda_{S,t}z_{S,t})$ and compute $\lambda_{S,t+1}$ from $z_{S,1},\ldots,z_{S,t}$ via~\eqref{eq:ftrl-opt}.
  \ENDFOR
  \STATE Update the sample mean $\widehat\mu_{t,i}$ for each $i\in Q_t$ from its observed scores, leaving unqueried entries unchanged.
  \STATE Set $\mathcal A_t\gets\mathcal A_{t-1}$; delete $S\in\mathcal A_t$ in order when $E_{S,t}\ge1/\alpha$, stopping once $|\mathcal A_t|=1$. If $|\mathcal A_t|=1$, \textbf{return} the remaining set.
\ENDFOR
\end{algorithmic}
\end{algorithm}

\subsection{Target Qualification}\label{sec:support-query}

M4 determines whether $J$ qualifies for the Top-$K$ set, so it queries $J$ on every question. As in M3, on question $t$ it queries all models with probability $p_t=t^{-1/2}$; otherwise, it chooses opponents using the rules below, based only on observations before question $t$.

\subsubsection{Estimating Score Differences}

The target answers every question, but each opponent may answer only a subset. We therefore average each opponent's score differences with $J$ over their shared questions, canceling any additive question effect common to the pair.

For opponent $i$, let $N_i=\sum_{s<t}\mathbf1\{i\in Q_s\}$ be its number of observations before question $t$. Set $\widehat\mu_J=0$ and define
\[
 \widehat\mu_i=
 \begin{cases}
 \displaystyle\frac{1}{N_i}\sum_{s<t:\,i\in Q_s}(X_{s,i}-X_{s,J}),&N_i>0,\\[6pt]
 0,&N_i=0.
 \end{cases}
\]
Here $\widehat\mu_i$ estimates $\mu_i-\mu_J$ rather than $\mu_i$, and is positive when $i$ has scored higher than $J$ on average over their shared questions. Since $C_A^+$ and $C_B^-$ are unchanged by adding a constant to every coordinate, we project $\widehat\mu$ directly onto them.

\Needspace{5\baselineskip}
\subsubsection{Choosing a Hypothesis Family}

We use the paired estimates to decide whether to focus on certifying IN or OUT. On a selective query, let $\widehat\mu_{(K)}$ be the $K$-th largest estimated difference among the opponents. Its sign determines which family to focus on:
\begin{itemize}[leftmargin=1.5em]
 \item If $\widehat\mu_{(K)}<0$, the estimates favor IN. Select the unrejected plus hypotheses in $\mathcal A_{t-1}^+$, since rejecting all of them certifies IN.
 \item If $\widehat\mu_{(K)}>0$, the estimates favor OUT. Select the unrejected minus hypotheses in $\mathcal A_{t-1}^-$, since rejecting all of them certifies OUT.
 \item If $\widehat\mu_{(K)}=0$, query all models by setting $Q_t=[M]$.
\end{itemize}

\subsubsection{Choosing Opponents}

When $\widehat\mu_{(K)}\ne0$, start with $Q_t=\{J\}$. For each unrejected hypothesis in the selected family, add to $Q_t$ every opponent selected by either of the following two rules.

First, include every opponent with a nonzero payoff weight. For a hypothesis cone $C$, project the full vector of past estimates to obtain $r=\widehat\mu-\Pi_C\widehat\mu$, and add every opponent $i$ with $r_i\ne0$. Any opponent with $r_i=0$ has zero weight in~\eqref{eq:projection-payoff}, so its score is not needed to evaluate that hypothesis's current payoff.

Second, include opponents that could still be the weakest in $A$ for $H_A^+$ or the strongest in $B$ for $H_B^-$ once estimation noise is considered, even when $r_i=0$, since those extremes best refute each hypothesis.

Compute the standard error of each paired mean from observations before question $t$:
\[
\widehat{\mathrm{SE}}_i=
\begin{cases}
\displaystyle
\sqrt{
\tfrac{
\sum_{s<t}\mathbf1\{i\in Q_s\}
\left(X_{s,i}-X_{s,J}-\widehat\mu_i\right)^2
}{
N_i(N_i-1)
}},
&N_i\ge2,\\[8pt]
1,&N_i<2.
\end{cases}
\]

For a plus hypothesis $H_A^+$, add each $i\in A$ satisfying
\[
 \widehat\mu_i-2\widehat{\mathrm{SE}}_i
 <\min_{j\in A}\widehat\mu_j,
\]
and for a minus hypothesis $H_B^-$, add each $i\in B$ satisfying
\[
 \widehat\mu_i+2\widehat{\mathrm{SE}}_i
 >\max_{j\in B}\widehat\mu_j.
\]
These rules choose which models to query; they are not confidence bounds used to reject hypotheses. The e-processes provide the error guarantee.

Because $\widehat{\mathrm{SE}}_i=1$ when $N_i<2$, any opponent $i$ with $N_i<2$ in an unrejected hypothesis of the selected family automatically satisfies this condition and is queried.

\Needspace{5\baselineskip}
\subsubsection{Updating Evidence and Stopping}

Before observing scores on $Q_t$, compute a residual for every unrejected hypothesis in both families by restricting $\widehat\mu$ and its cone to $Q_t$ and projecting onto the restricted cone. Once the fresh scores arrive, use this residual to compute the payoff in~\eqref{eq:projection-payoff}, summing only over $Q_t$ and using zero when the residual is zero. Update every wealth account, choose its next betting coefficient by~\eqref{eq:ftrl-opt}, and update the paired estimates from the fresh observations. Apply the deletion and stopping rule in Section~\ref{sec:membership-output}. While the selected family determines which opponents to query, both families continue to accumulate evidence. These steps define M4.

For each hypothesis in the selected family, $Q_t$ contains all coordinates with nonzero residual in its full projection. Restricting to $Q_t$ therefore preserves its residual weights and its current payoff. Appendix~\ref{app:query-preservation} gives the proof. Algorithm~\ref{alg:pace-m4} summarizes M4.

\begin{algorithm}[htbp]
\caption{Selective Target Qualification: M4}
\label{alg:pace-m4}
\small
\begin{algorithmic}[1]
\REQUIRE Models $[M]$, cutoff $K$, target $J$, error level $\alpha$.
\STATE Initialize $\mathcal A_0^+$ and $\mathcal A_0^-$ as in Algorithm~\ref{alg:pace-full}, setting $E_{C,0}\gets1$ and $\lambda_{C,1}\gets0$ for each cone $C$.
\STATE Set $\widehat\mu_J\gets0$, and for each $i\ne J$ set $N_i\gets0$, $\widehat\mu_i\gets0$, and $\widehat{\mathrm{SE}}_i\gets1$.
\FOR{$t=1,2,\ldots$}
  \STATE Let $\widehat\mu_{(K)}$ be the $K$-th largest entry of $(\widehat\mu_i)_{i\ne J}$.
  \IF{a full query is drawn with probability $p_t=t^{-1/2}$ \OR $\widehat\mu_{(K)}=0$}
    \STATE Set $Q_t\gets[M]$.
  \ELSE
    \STATE Select $\mathcal A_{t-1}^+$ if $\widehat\mu_{(K)}<0$, or $\mathcal A_{t-1}^-$ if $\widehat\mu_{(K)}>0$, and initialize $Q_t\gets\{J\}$.
    \FOR{each cone $C$ in the selected family}
      \STATE Compute $r\gets\widehat\mu-\Pi_C\widehat\mu$ and add every $i\ne J$ with $r_i\ne0$ to $Q_t$.
      \STATE If $C=C_A^+$, add each $i\in A$ with $\widehat\mu_i-2\widehat{\mathrm{SE}}_i<\min_{j\in A}\widehat\mu_j$ to $Q_t$; if $C=C_B^-$, add each $i\in B$ with $\widehat\mu_i+2\widehat{\mathrm{SE}}_i>\max_{j\in B}\widehat\mu_j$ to $Q_t$.
    \ENDFOR
  \ENDIF
  \STATE For every $C\in\mathcal A_{t-1}^+\cup\mathcal A_{t-1}^-$, restrict $\widehat\mu$ and $C$ to $Q_t$ and project to obtain residual $r_C$.
  \STATE Query $Q_t$ and observe $X_{t,Q_t}=(X_{t,i})_{i\in Q_t}$.
  \FOR{every $C\in\mathcal A_{t-1}^+\cup\mathcal A_{t-1}^-$}
    \STATE Set $z_{C,t}\gets 2r_C^\top X_{t,Q_t}/\|r_C\|_1$ if $r_C\ne0$, and $z_{C,t}\gets0$ otherwise.
    \STATE Update $E_{C,t}\gets E_{C,t-1}(1+\lambda_{C,t}z_{C,t})$ and compute $\lambda_{C,t+1}$ from $z_{C,1},\ldots,z_{C,t}$ via~\eqref{eq:ftrl-opt}.
  \ENDFOR
  \STATE For each $i\in Q_t\setminus\{J\}$, increment $N_i$ and update $\widehat\mu_i$ and $\widehat{\mathrm{SE}}_i$ from $(X_{s,i}-X_{s,J})_{s\le t:\,i\in Q_s}$, keeping $\widehat{\mathrm{SE}}_i=1$ when $N_i<2$.
  \STATE Set $\mathcal A_t^+\gets\mathcal A_{t-1}^+$ and $\mathcal A_t^-\gets\mathcal A_{t-1}^-$; delete $C$ in order when $E_{C,t}\ge1/\alpha$, stopping once either family is empty. \textbf{Return} $\mathrm{IN}$ if $\mathcal A_t^+=\varnothing$, or $\mathrm{OUT}$ if $\mathcal A_t^-=\varnothing$.
\ENDFOR
\end{algorithmic}
\end{algorithm}

\section{Theoretical Guarantees}\label{sec:theory}

We establish three types of guarantees. First, M1--M4 control the probability of an incorrect answer, even though the stopping time depends on the data. Second, they return an answer with finite expected query cost when the relevant mean gap is positive, and a positive gap is necessary for almost sure termination. Finally, we compare query costs: selective versus full observation for the same task, and direct target qualification versus identifying the entire Top-$K$ set.

\subsection{Anytime Validity}

To state a common guarantee for M1--M4, write $y^\star$ for the true answer: $S^\star$ for M1 and M3, or $\ell^\star$ for M2 and M4. The set of answers still possible after question $t$ is
\[
\mathcal Y_t=
\begin{cases}
\mathcal A_t,&\text{for M1 and M3},\\[4pt]
\{\mathrm{IN}:\mathcal A_t^-\ne\varnothing\}\cup\{\mathrm{OUT}:\mathcal A_t^+\ne\varnothing\},&\text{for M2 and M4}.
\end{cases}
\]

Each method returns when exactly one answer remains. Its first output time is
\begin{equation}\label{eq:first-output}
 \tau=\inf\{t\ge1:|\mathcal Y_t|=1\}.
\end{equation}
When $\tau<\infty$, the output is the sole retained answer, $\mathcal Y_\tau=\{\widehat y_\tau\}$.

\begin{theorem}\label{thm:valid}\label{cor:first-output}
For each method M1--M4, the retained answers satisfy
\[
 \Prb\{\exists t\ge0:y^\star\notin\mathcal Y_t\}\le\alpha.
\]
The first output in~\eqref{eq:first-output} satisfies $\Prb\{\tau<\infty,\ \widehat y_\tau\ne y^\star\}\le\alpha$. Thus, the probability of returning an incorrect answer is at most $\alpha$.
\end{theorem}

\subsection{Stopping and Expected Queries}

To ensure an answer, we require a positive gap between the means that determine the decision. For Top-$K$ identification, this is the boundary gap $\Delta_K$. For target qualification, define $\delta_J$ as the absolute difference between $\mu_J$ and the $K$-th largest mean among its opponents. The target gap can remain positive even when other models tie at the Top-$K$ boundary and $\Delta_K=0$.

\begin{theorem}\label{thm:stopping}
Assume $\Delta_K>0$ for M1 and M3, and $\delta_J>0$ for M2 and M4. There are constants $A,c,t_0>0$ such that $\Prb(\tau>t)\le Ae^{-c\sqrt t}$ for all integers $t\ge t_0$. Consequently, $\E\{\tau+\Cost_\tau\}<\infty$.
\end{theorem}

Thus, each method returns an answer with probability one, and both the expected number of questions and the expected number of model queries are finite.

The positive-gap condition is also necessary for almost sure termination.

\begin{proposition}\label{prop:zero-gap}
Fix $\alpha\in(0,1/2)$ and either task. Suppose a sequential procedure has error probability at most $\alpha$ under every bounded i.i.d.\ score distribution with a positive gap. Then under every bounded i.i.d.\ score distribution $P$ with zero gap, its stopping time $\tau$ satisfies
\[
 \Prb_P(\tau<\infty)\le 2\alpha,\qquad \Prb_P(\tau=\infty)\ge 1-2\alpha.
\]
\end{proposition}

Because $1-2\alpha>0$, no valid procedure can terminate almost surely when the relevant gap is zero.

\subsection{Selective Query Savings}\label{sec:adaptive-comparison}

We now compare selective and full observation for the same task. Let $\tau_j$ be M$j$'s first output time and $C_j=\Cost_{\tau_j}$ its total model queries, for $j=1,\ldots,4$. The comparisons are M3 versus M1 for Top-$K$ identification, and M4 versus M2 for target qualification. Fix $\alpha\in(0,1/2)$, $M\ge3$, $K$, and positive constants $\gamma,v_0$. The results below consider one increasingly close pair while all other relevant mean differences remain separated by at least $\gamma$.

\subsubsection{Top-\texorpdfstring{$K$}{K} Identification}
For set identification, we compare M3 with M1 when the main difficulty is distinguishing the models at ranks $K$ and $K+1$. Every other model has a mean at least $\gamma$ above both, or at least $\gamma$ below both. Assume that the difference between the two boundary models' scores on the same question has variance at least $v_0$. Thus the difficult comparison retains random variation as its mean gap shrinks.

\begin{theorem}\label{thm:adaptive-set}
As $\Delta_K\downarrow0$, $\E\tau_j=\Theta(\Delta_K^{-2}\log(1/\Delta_K))$ for $j\in\{1,3\}$, and $\E C_3/\E\tau_3\to2$. In the same limit,
\[
 \limsup_{\Delta_K\downarrow0}\frac{\E\tau_3}{\E\tau_1}
 \le\frac{1}{1-2\alpha},\quad
 \limsup_{\Delta_K\downarrow0}\frac{\E C_3}{\E C_1}
 \le\frac{2}{M(1-2\alpha)}.
\]
\end{theorem}

When the main difficulty is distinguishing the models at ranks $K$ and $K+1$, M3 and M1 require the same order of expected questions. For M3, expected total queries approach twice the expected number of questions, whereas M1 queries all $M$ models on every question.

\Needspace{5\baselineskip}
\subsubsection{Target Qualification}
For target qualification, we make the analogous comparison between M4 and M2, with the difficulty now arising between the target and its cutoff opponent. Among the opponents of $J$, let $u$ have the $K$-th largest mean. If $\mu_J>\mu_u$, the target is IN; if $\mu_J<\mu_u$, it is OUT. Assume every other opponent has a mean at least $\gamma$ above both $J$ and $u$, or at least $\gamma$ below both. Also assume that the difference between $J$'s and $u$'s scores on the same question has variance at least $v_0$.

\begin{theorem}\label{thm:adaptive-membership}
As $\delta_J\downarrow0$, $\E\tau_j=\Theta(\delta_J^{-2}\log(1/\delta_J))$ for $j\in\{2,4\}$, and $\E C_4/\E\tau_4\to2$. In the same limit,
\[
 \limsup_{\delta_J\downarrow0}\frac{\E\tau_4}{\E\tau_2}
 \le\frac{1}{1-2\alpha},\quad
 \limsup_{\delta_J\downarrow0}\frac{\E C_4}{\E C_2}
 \le\frac{2}{M(1-2\alpha)}.
\]
\end{theorem}

Similar to Theorem~\ref{thm:adaptive-set}, when $J$ and $u$ are close, M4 matches M2 in the order of expected questions while reducing the number of queries per question from $M$ to $2$.

\subsection{Target vs.\ Set Identification}

The preceding results held the task fixed and changed which models were queried. We now ask whether recovering the entire Top-$K$ set is necessary when only $J$'s qualification matters. Suppose $J$ is clearly above or below its cutoff, while two other models at ranks $K$ and $K+1$ have nearly equal means. Identifying the set requires distinguishing that pair. Their order does not change whether $J$ qualifies, so direct qualification can avoid this difficult comparison.

Fix $\alpha\in(0,1/2)$. Consider any Top-$K$ identification procedure that, under every bounded i.i.d.\ score distribution with a strict Top-$K$ boundary, returns an answer with probability one and has first-output error probability at most $\alpha$. Let $\tau$ be its first output time, and let $\operatorname{kl}(p,q)$ denote binary relative entropy.

\begin{theorem}\label{thm:separation}
For every fixed $\delta\in(0,1]$, there is a finite $B=B(M,K,\alpha,\delta)$ such that $\sup_{\delta_J\ge\delta}\E C_j\le B$ for $j\in\{2,4\}$. Every Top-$K$ identification procedure described above satisfies, whenever $\Delta_K>0$,
\[
 \E\Cost_\tau
 \ge\frac{\operatorname{kl}(1-\alpha,\alpha)}{\Delta_K}.
\]
For full observation, the right-hand side is multiplied by $M$.
\end{theorem}

Along distributions with a fixed positive target gap and a shrinking Top-$K$ boundary gap, direct qualification stays bounded in expected cost while complete set identification becomes arbitrarily costly.

\section{Numerical Experiments}\label{sec:experiments}

We evaluate Top-$K$ identification and target qualification on synthetic scores and six Open LLM Leaderboard benchmarks at nominal error level $\alpha=0.05$. Within each comparison, the methods use the same i.i.d.\ streams of question score vectors. Across all settings, empirical first-output error rates for PACE remained within $\alpha$; see Appendix Table~\ref{tab:pace-first-output-errors}.

\begin{itemize}[leftmargin=1.5em,itemsep=2pt]
  \item Under full observation, all $M$ models answer every question. We compare M1 (Top-$K$ identification) and M2 (target qualification) with RankCS~\cite{khosravi2026rank} and PrPl-EB~\cite{10.1093/jrsssb/qkad009} by the mean number of questions to the first answer.
  \item Under selective querying, we compare the mean total number of model queries until an answer is returned. One model answering one question counts as one query. We compare M3 (Top-$K$ identification) and M4 (target qualification) with RankCS using Top-$K$ retirement and with the racing methods BRR and VRR~\cite{gharat2025representative}. For identification, we also compare M3 with KL-LUCB~\cite{kaufmann2013information} and LUCB++~\cite{simchowitz2017simulator}. We use the same metric to compare M3 with M1 and M4 with M2, which measures the savings from selective querying within PACE. On the leaderboard benchmarks we also compare M3 and M4 with PrPl-EB by model queries.
\end{itemize}

With or without retirement, RankCS uses the default five-point mixture bet from the original paper~\cite{khosravi2026rank}. Appendix~\ref{app:rankcs-ftrl} repeats the comparisons with RankCS using the PACE FTRL rule~\eqref{eq:ftrl-opt} for each model pair. The retirement rule stops querying a model once its rank bounds certify it inside or outside the Top-$K$. For target qualification, RankCS stops as soon as the surviving rankings agree on the target's membership. BRR and VRR also stop once they determine whether the target qualifies.

In the synthetic experiments, model $i$ answers each question correctly with probability $\mu_i$. Index the models so that $\mu_1\ge\cdots\ge\mu_M$, so the Top-$K$ boundary gap is $\Delta_K=\mu_K-\mu_{K+1}$.

To vary how models perform on the same question while keeping their accuracies fixed, we generate binary scores with a Gaussian copula. For each question $t$, we independently draw a shared difficulty $Z_t$ and model-specific terms $\varepsilon_{t,1},\ldots,\varepsilon_{t,M}$, all standard normal. Model $i$ scores
\[
 X_{t,i}=\mathbf{1}\!\left\{\sqrt{\rho}\,Z_t+\sqrt{1-\rho}\,\varepsilon_{t,i}\leq\Phi^{-1}(\mu_i)\right\},
\]
where $\Phi$ is the standard normal distribution function, so $\Prb(X_{t,i}=1)=\mu_i$.

\subsection{Within-Question Score Correlation}\label{sec:exp-e2}

\begin{figure}[htbp]
\centering
\includegraphics[width=0.90\textwidth]{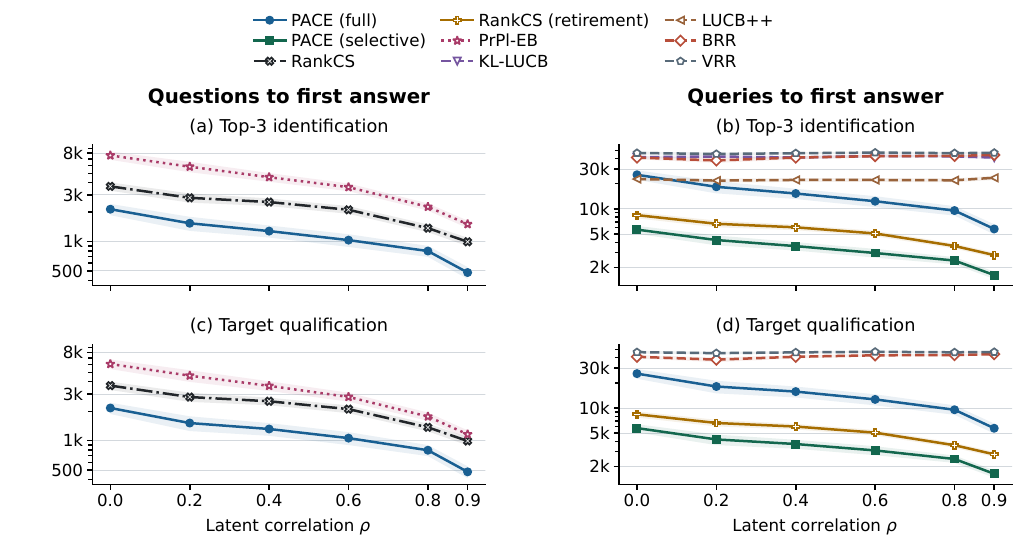}
\caption{Mean questions under full observation (left) and mean total model queries under selective querying (right) as $\rho$ increases, with $K=3$ and $\Delta_K=0.05$. Full-observation PACE uses M1 for the set and M2 for the target. Selective PACE uses M3 and M4, respectively. Bands show $\pm 2$ standard errors over 100 paired streams.}
\label{fig:e2-correlation}
\end{figure}

Variation in question difficulty can induce positive correlation between the scores of different models on the same question. With each model's accuracy held fixed, how does the number of queries PACE needs change as this positive correlation grows stronger? We use 12 models with $K=3$ and boundary gap $\Delta_K=0.05$: the models at ranks $K$ and $K+1$ have accuracies $0.5+\Delta_K/2$ and $0.5-\Delta_K/2$, those above them have accuracy $0.75$, and those below them have accuracy $0.25$. We identify the Top-3 set or test whether the model at rank 3 qualifies. We vary $\rho$ from $0$ to $0.9$, with all methods evaluated on the same 100 streams at each level. A larger $\rho$ induces stronger positive dependence. Figure~\ref{fig:e2-correlation} shows the resulting query costs.

Full-observation PACE uses fewer questions than RankCS and PrPl-EB at every correlation level shown. As $\rho$ rises from $0$ to $0.9$, the mean falls from $2{,}138$ to $480$ questions for M1 and from $2{,}158$ to $479$ for M2. At $\rho=0.9$, full-observation RankCS needs $992$ questions on each task, so M1 and M2 use $51.6\%$ and $51.7\%$ fewer questions, respectively. Stronger correlation also reduces query costs under selective querying. For Top-3 identification, M3 uses $5{,}659$ queries on average at $\rho=0$ and $1{,}616$ at $\rho=0.9$, compared with $8{,}439$ and $2{,}804$ for RankCS with retirement, reductions of $32.9\%$ and $42.4\%$, respectively. For qualifying the rank-3 target, M4 uses $5{,}770$ queries on average at $\rho=0$ and $1{,}637$ at $\rho=0.9$, compared with $8{,}436$ and $2{,}800$ for RankCS with retirement ($31.6\%$ and $41.5\%$ fewer). Relative to their full-observation counterparts M1 and M2 at $\rho=0.9$, M3 and M4 save $72.0\%$ and $71.5\%$ of model queries.

The costs of KL-LUCB, LUCB++, BRR, and VRR change little across $\rho$, because these methods do not cancel shared question difficulty. Figure~\ref{fig:e2-correlation} considers model scores that are independent or positively correlated on the same question; when scores are negatively correlated across the cutoff, PACE still uses fewer questions and queries than all baselines (Appendix Table~\ref{tab:opposed-costs}).

\subsection{Boundary Gap and Pool Size}\label{sec:exp-e3}

Using the same accuracy profile as in Section~\ref{sec:exp-e2}, we vary the boundary gap $\Delta_K$ and the number of models $M$ in separate experiments, each with $K=3$ and $\rho=0.7$. At $M=10$, we decrease $\Delta_K$ through $0.12$, $0.08$, $0.05$, $0.03$, $0.02$, and $0.015$. At $\Delta_K=0.05$, we increase $M$ through $6$, $12$, $20$, $40$, and $60$ by adding models of mean $0.25$. At each setting we identify the Top-3 set and test qualification of the model at rank 3 on the same 100 streams.

\begin{figure}[htbp]
\centering
\includegraphics[width=0.90\textwidth]{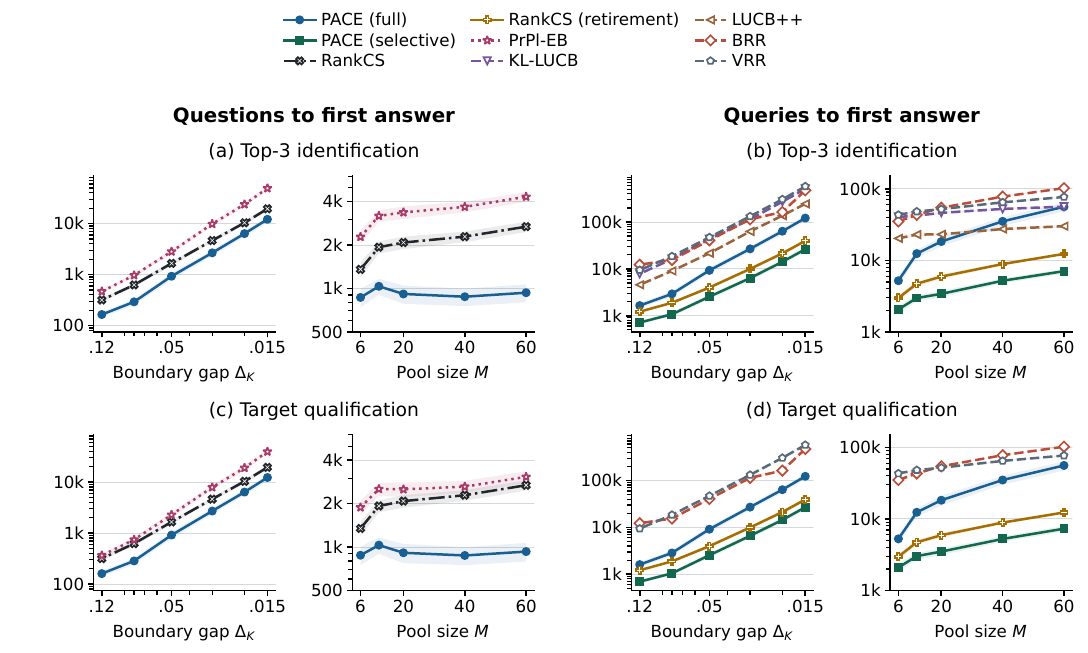}
\caption{Scaling with boundary gap and pool size at $K=3$ and $\rho=0.7$. The left column compares full-observation methods by mean questions. The right column compares mean total model queries. Within each panel, the left plot varies $\Delta_K$ at $M=10$, and the right plot varies $M$ at $\Delta_K=0.05$. Bands show $\pm 2$ standard errors over 100 paired streams per setting.}
\label{fig:e3-scaling}
\end{figure}

As the boundary gap shrinks from $0.12$ to $0.015$ at $M=10$, mean queries for Top-3 identification rise from $716$ to $26{,}156$ for M3 and from $1{,}218$ to $39{,}537$ for RankCS with retirement, so M3 uses $41.3\%$ and $33.8\%$ fewer queries at the two endpoints. For qualifying the rank-3 target over the same gaps, M4 rises from $694$ to $26{,}945$ queries versus $1{,}218$ to $39{,}395$ for RankCS with retirement ($43.0\%$ and $31.6\%$ fewer). Across these gaps, the ratio of mean total queries to mean questions falls from $4.35$ to $2.16$ for M3 and from $4.26$ to $2.18$ for M4, approaching $2$ as predicted by Theorems~\ref{thm:adaptive-set} and~\ref{thm:adaptive-membership}. At $M=60$ and $\Delta_K=0.05$, full-observation PACE uses $935$ questions for identification and $930$ for qualification on average, compared with $2{,}681$ for RankCS ($65.1\%$ and $65.3\%$ fewer). With retirement, RankCS uses about $12{,}308$ model queries on each task, while M3 and M4 use $7{,}065$ and $7{,}308$ ($42.6\%$ and $40.6\%$ fewer). Relative to M1 and M2, M3 and M4 save $87.4\%$ and $86.9\%$ of model queries. Adding models far from the cutoff forces full-observation methods to query all $M$ models on every question, whereas M3 and M4 focus on models near the cutoff. Their savings over full observation therefore grow with $M$.

\subsection{Target vs.\ Set Identification}\label{sec:exp-e4}

\begin{figure}[htbp]
\centering
\includegraphics[width=0.90\textwidth]{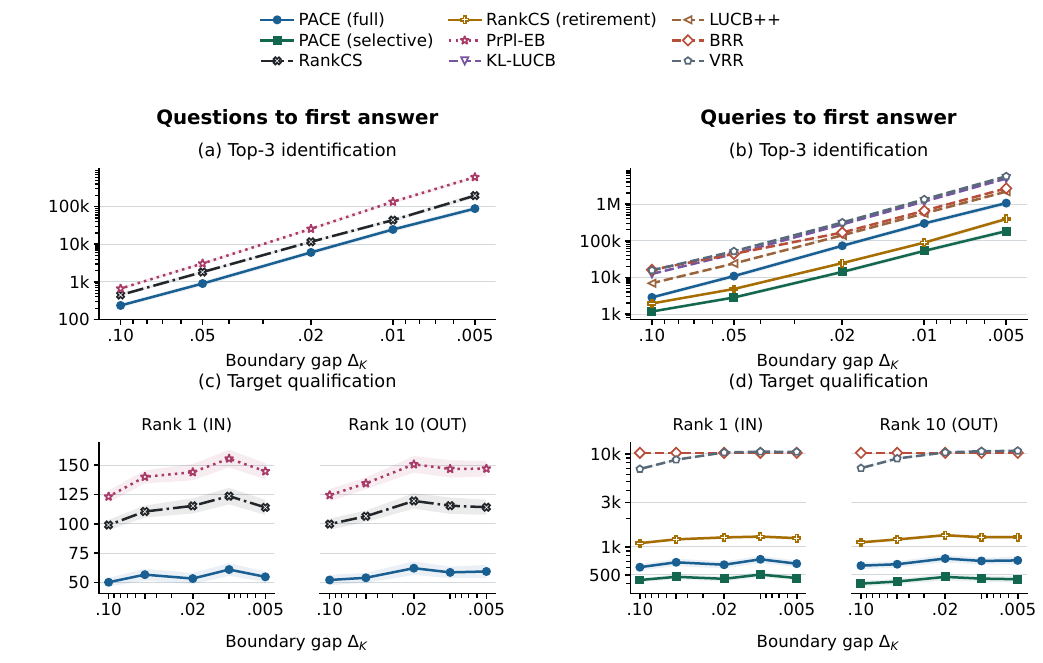}
\caption{Mean questions under full observation (left) and mean total model queries (right) as the Top-3 boundary gap $\Delta_K$ shrinks at $\rho=0.7$. The bottom row tests targets at ranks 1 and 10. Vertical axes are logarithmic except in (c). Bands show $\pm 2$ standard errors over 100 paired streams per gap.}
\label{fig:e4-separation}
\end{figure}

If the target is far from the cutoff, does a near-tie at the cutoff make its qualification more costly? To examine the separation in Theorem~\ref{thm:separation}, we use $M=12$ models, $K=3$, and $\rho=0.7$. The model accuracies are $(0.70,0.65,0.50+\Delta_K/2,0.50-\Delta_K/2,0.30,\ldots,0.30)$. We shrink $\Delta_K$ from $0.10$ to $0.005$, varying only the accuracies at ranks 3 and 4. On the same 100 paired streams per gap, we compare recovering the Top-3 set with deciding membership for a clearly IN target at rank 1 or a clearly OUT target at rank 10. Figure~\ref{fig:e4-separation} shows the mean numbers of questions and model queries to the first answer.

A near-tie at the Top-3 cutoff makes full-set identification expensive, but does not affect well-separated targets. As $\Delta_K$ shrinks from $0.10$ to $0.005$, mean queries for Top-3 identification rise from $1{,}170$ to $182{,}667$ under M3, and M1 requires $88{,}362$ questions on average at $\Delta_K=0.005$. For the rank-1 and rank-10 targets, cost stays nearly flat across the five gaps: M4 uses $401$ to $501$ queries on average ($460$ for rank 1 and $447$ for rank 10 at $\Delta_K=0.005$), and M2 needs $55$ and $59$ questions, settling both targets without waiting to distinguish ranks 3 and 4.

Across both tasks, PACE also uses far fewer questions and queries than RankCS and the other baselines in Figure~\ref{fig:e4-separation}. At $\Delta_K=0.005$, full-observation RankCS needs $195{,}374$ questions for Top-3 identification and about $114$ for either target; M1 and M2 reduce mean questions by $54.8\%$ for the set, $51.9\%$ for rank 1, and $48.0\%$ for rank 10. Under selective querying at $\Delta_K=0.005$, RankCS with retirement uses $391{,}279$ queries for the set and $1{,}237$ and $1{,}271$ for ranks 1 and 10, whereas M3 and M4 reduce mean total queries by $53.3\%$, $62.8\%$, and $64.8\%$, respectively.

\subsection{Open LLM Leaderboard Evaluation}\label{sec:exp-e6}

Finally, we evaluate PACE on real per-item scores of 395 models from the Open LLM Leaderboard via tinyBenchmarks~\cite{polo2024tinybenchmarks} across six benchmarks: MMLU, HellaSwag, GSM8K, TruthfulQA, WinoGrande, and ARC-Challenge. On each benchmark, the ground-truth mean is the average over all items, and each step draws one item uniformly with replacement to form an i.i.d.\ stream preserving real question difficulty and cross-model correlations.

\begin{figure}[htbp]
\centering
\includegraphics[width=\linewidth]{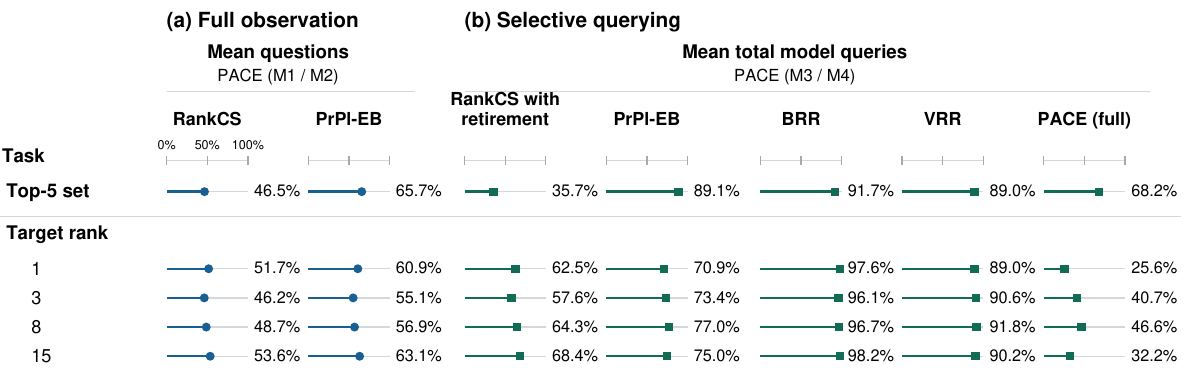}
\caption{Percentage reductions in mean questions (a) and mean total model queries (b) on GSM8K for Top-5 identification and qualification at the four target ranks. Each reduction compares means on the same 100 streams.}
\label{fig:e6-b15}
\end{figure}

On each benchmark, we select 15 models at evenly spaced rank quantiles across the 395 models and run 100 paired streams for Top-5 identification and target qualification at ranks $1, 3, 8$, and $15$. We cap each run at 100,000 questions. Figure~\ref{fig:e6-b15} shows GSM8K, the only benchmark on which every baseline returned a Top-5 set on all 100 streams. Appendix Figures~\ref{fig:app-leaderboard-costs}--\ref{fig:app-savings-arc-challenge} show the other five benchmarks. Appendix Table~\ref{tab:app-e6-b15-full} gives Top-5 identification query counts for all six benchmarks.

On GSM8K, M1 identifies the Top-5 set in $330$ questions on average, compared with $618$ for RankCS, a reduction of $46.5\%$. With selective querying, M3 uses $1{,}576$ model queries, compared with $2{,}451$ for RankCS with retirement, a reduction of $35.7\%$. M3 also uses $91.7\%$ fewer queries than BRR and $89.0\%$ fewer than VRR on this task.

\begin{figure}[!htbp]
\centering
\includegraphics[width=0.94\textwidth]{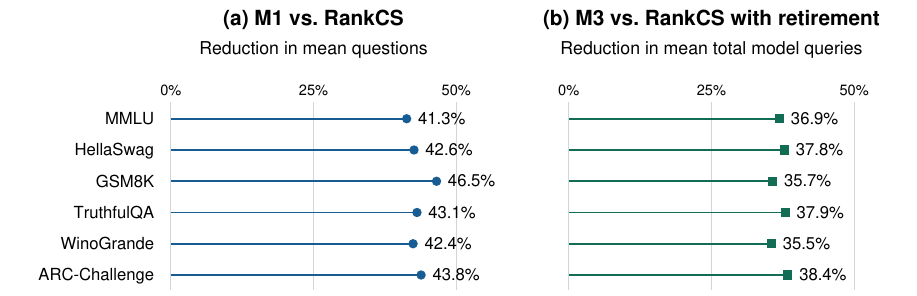}
\caption{Percentage reductions in mean questions (a) and mean total model queries (b) for Top-5 identification on each benchmark over 100 paired streams.}
\label{fig:leaderboard-query-savings}
\end{figure}

Figure~\ref{fig:leaderboard-query-savings} gives the Top-5 results for all six benchmarks. Under full observation, M1 uses $41.3\%$ to $46.5\%$ fewer questions than RankCS. Under selective querying, M3 uses $35.5\%$ to $38.4\%$ fewer model queries than RankCS with retirement. For target qualification across the six benchmarks and four target ranks, M2 uses $34.6\%$ to $53.6\%$ fewer questions on average than RankCS, and M4 uses $45.6\%$ to $68.4\%$ fewer model queries than RankCS with retirement (Figure~\ref{fig:e6-b15} and Appendix Figures~\ref{fig:app-leaderboard-costs}--\ref{fig:app-savings-arc-challenge}).

\begin{figure}[htbp]
\centering
\includegraphics[width=0.90\textwidth]{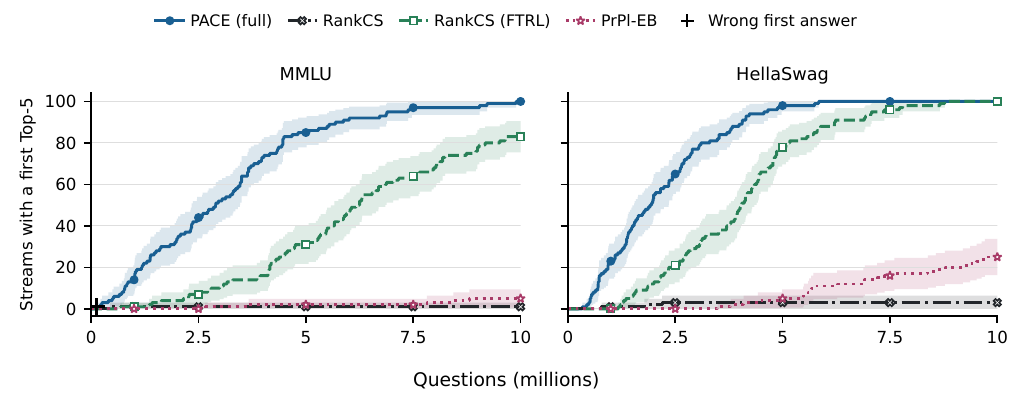}
\caption{Cumulative Top-5 returns over 100 paired streams per benchmark in the near-tied 20-model pools under full observation. The two RankCS curves use its default mixture and the PACE FTRL betting rule. Shading shows pointwise $\pm 2$ standard errors. Plus signs mark wrong first answers.}
\label{fig:e6-extended}
\end{figure}

On MMLU and HellaSwag we also evaluate the 20 highest-scoring models. Rank 5 leads rank 6 by only six items in each benchmark, with $\Delta_5=0.00043$ and $0.00060$, respectively. Figure~\ref{fig:e6-extended} shows when each method first identifies the Top-5 set. By the budget of $10^7$ questions, M1 returns on all 100 streams in each benchmark, with one wrong first answer on MMLU. With its default mixture, RankCS returns on $1/100$ MMLU streams and $3/100$ HellaSwag streams. Replacing the mixture with FTRL raises these counts to $83/100$ and $100/100$, respectively, with all returned sets correct. On HellaSwag, M1 and RankCS with FTRL use $2.14$ million and $4.04$ million questions on average, respectively. PrPl-EB returns on $5/100$ MMLU streams and $25/100$ HellaSwag streams.

\section{Discussion and Conclusion}

We have presented PACE, a sequential evaluation framework covering Top-$K$ identification and target qualification under both full observation and selective querying. PACE projects past mean estimates onto candidate cones to weight score differences on each shared question and updates e-processes via FTRL, canceling shared question effects while controlling finite-sample error at any stopping time. Across our synthetic and leaderboard experiments, M1 and M2 use fewer questions than full-observation baselines. Under selective querying, M3 and M4 focus on unresolved comparisons and reduce total model queries relative to both full observation and selective baselines.

Two limitations point to natural extensions. First, our theoretical guarantees assume i.i.d.\ bounded score vectors; extending PACE to non-stationary streams where the question distribution shifts over time remains open. Second, maintaining all candidate accounts can become expensive: Top-$K$ identification starts with $\binom{M}{K}$ accounts. An important next step is to represent or remove groups of candidates without listing every candidate separately.

{\small
\let\oldbibliography\thebibliography
\renewcommand{\thebibliography}[1]{%
  \oldbibliography{#1}%
  \setlength{\itemsep}{1.5pt}%
  \setlength{\parskip}{0pt}%
}
\bibliographystyle{plain}
\bibliography{reference}
}

\clearpage
\appendix
\section{Proofs}\label{app:proofs}

For the probability arguments, let $\mathcal F_n$ include all model scores and the independent random numbers used to choose queries through row $n$. This includes scores that the algorithm did not query; the algorithm itself still uses only observed scores. The next row $X_{n+1}$ is independent of this history and of the next query random number $U_{n+1}$.

\subsection{Proof of Theorem~\ref{thm:valid}}\label{app:validity}

Given the history and current query uniform, the query and projection direction are fixed. Under a true hypothesis, $\mu_{Q_t}$ belongs to the restricted cone $C$. The empirical projection residual $r$ lies in its polar cone, so $r^\top\mu_{Q_t}\le0$. Since $C$ contains the constant vectors, $r^\top\mathbf1=0$. For $r\ne0$, the positive and negative weights in $2r/\|r\|_1$ each have total magnitude one. Hence the payoff lies in $[-1,1]$ and has nonpositive conditional mean, also when $r=0$ and the payoff is zero. The wealth stopped at rejection or first output is a nonnegative supermartingale starting at one. By Ville's inequality, the probability of ever rejecting this hypothesis is at most $\alpha$.

For M1 and M3, take $H_{S^\star}$. For M2 and M4, fix $M-K$ opponents with $\mu_i\le\mu_J$ if the target is IN, or $K$ opponents with $\mu_i>\mu_J$ if it is OUT. Their hypothesis ($H_B^-$ or $H_A^+$, respectively) is true. In each case, losing the true answer requires rejecting the chosen hypothesis, as does an incorrect first output. The retained set stays fixed after output.

\subsection{Proof of Theorem~\ref{thm:stopping}}

If the procedure has not returned an answer, at least one false hypothesis remains. We first bound how long its FTRL wealth takes to reach the rejection threshold, allowing any initial sequence of scores followed by scores with positive conditional means. We then show that accurate estimates provide this positive mean under either query rule.

\subsubsection{Crossing a Wealth Threshold}\label{app:ftrl-update}

For scores $z_j\in[-1,1]$, let $b_n$ maximize the FTRL objective~\eqref{eq:ftrl-opt} after $n$ scores, with penalty $R(b)=-\tfrac12\log(1-b^2)$. Write $W_n=\prod_{j\le n}(1+b_{j-1}z_j)$ and $T=\inf\{n:\log W_n\ge h\}$ for a fixed $h>0$.

\begin{lemma}\label{lem:aq-scalar-crossing}
For every bounded score path and every $n\ge0$,
\begin{equation}\label{eq:aq-scalar-crossing}
 \log W_n\ge
 \max_{0\le b\le1/2}\left\{\sum_{j\le n}\log(1+bz_j)-R(b)\right\}
 -16\log\left(1+\sum_{j\le n}z_j^2\right).
\end{equation}
If the first $s$ scores are arbitrary and subsequent scores have
conditional means at least $d>0$, then, for a constant $A_h$,
\begin{equation}\label{eq:aq-scalar-tail}
 \Prb(T>s+n)\le e^{-d^2n/512}
 \quad\text{whenever}\quad
 nd^2\ge A_h\{ds+\log(1+s+n)+1\}.
\end{equation}
\end{lemma}

\begin{proof}
Let $V_n=\sum_{j\le n}z_j^2$ and denote the regularized maximum in
\eqref{eq:aq-scalar-crossing} by $\Phi_n$. The bounds
$|z/(1+bz)|\le2|z|$, $z^2/(1+bz)^2\ge\kappa z^2$ and
$R''(b)\ge1$ hold throughout the coefficient range, with $\kappa=4/9$.
Strong concavity and the endpoint optimality conditions imply
\[
 |b_n-b_{n-1}|\le\frac{2|z_n|}{1+\kappa V_{n-1}},\qquad
 0\le\Phi_n-\Phi_{n-1}-\log(1+b_{n-1}z_n)
 \le\frac{4z_n^2}{1+\kappa V_{n-1}}.
\]
Since $V_n-V_{n-1}\le1$, integration of $(1+\kappa v)^{-1}$ gives
\[
 0\le\Phi_n-\log W_n
 \le\frac{4(1+\kappa)}{\kappa}\log(1+\kappa V_n)
 \le16\log(1+V_n).
\]
This proves~\eqref{eq:aq-scalar-crossing} for every history and time.

For~\eqref{eq:aq-scalar-tail}, compare with the fixed coefficient
$\beta=d/2$. Since $\log(1+u)\ge u-u^2$ on $[-1/2,1/2]$, the first
$s$ comparator returns sum to at least $-ds$, and each subsequent
conditional mean is at least $d^2/4$. On $\{T>s+n\}$,
\eqref{eq:aq-scalar-crossing} bounds their total above by
$h+R(\beta)+16\log(1+s+n)$.
Choose $A_h$ so that the difference from the lower conditional mean
is at least $nd^2/8$. Each centered comparator increment has absolute
value at most $2d$, because the derivative of $\log(1+\beta z)$
with respect to $z$ has absolute value at most $d$.
Azuma's inequality gives~\eqref{eq:aq-scalar-tail}.
\end{proof}

\subsubsection{Query Preservation}\label{app:query-preservation}

For M3, when $\widehat S_{t-1}=S^\star$, every ordinary query contains $S^\star\mathbin\triangle S$ for each retained set $S$. If $S$ is false, it omits some $i\in S^\star$ and includes some $j\notin S^\star$. Thus every query contains a pair violating $S$'s order by $\mu_i-\mu_j\ge\Delta_K$, even if the surviving family changes.

For M4, we show that the query preserves the full projection payoff for each hypothesis in the selected family. Let $A$ be its opponent set, $s=+1$ for a plus hypothesis and $s=-1$ for a minus hypothesis. At target coordinate zero, put $v_i=-s\widehat\mu_i$. In signed coordinates, projection fits a common threshold $h$ to the target and truncates each $v_i$ above $h$, minimizing $h^2+\sum_{i\in A}(v_i-h)_+^2$. Its unique minimizer solves
\[
 h=\sum_{i\in A}(v_i-h)_+.
\]
If $h>0$, normalization gives $a_J=s$ and $a_i=-s(v_i-h)_+/h$. Every query includes $J$ and all opponents with $v_i>h$. Removing any other opponents leaves the root and these coefficients unchanged. If $h=0$, every constraint already holds and the restricted residual is also zero.

\subsubsection{Accuracy Under Predictable Sampling}\label{app:sampling-accuracy}

We need to control estimation error at every row after a fixed starting row. For this proof only, continue with full queries after output so that all later estimates are defined. Each coordinate or jointly observed target--opponent difference is selected before the new row is observed. Its successive observations are therefore iid with the corresponding marginal law. For M3/M4 the independent full explorations through row $s$ have total expected count $\sum_{t\le s}t^{-1/2}\ge\sqrt s$. By Chernoff's inequality their count is at least $\sqrt s/2$ except with probability $e^{-\sqrt s/8}$. M1/M2 observe every coordinate $s$ times by that row.

For fixed $\epsilon>0$, apply Hoeffding's inequality to each local sample count $m\ge\sqrt s/2$ and sum over $m$ and at most $M$ coordinates. All observations lie in $[-1,1]$. Thus, in any of the four methods, the probability that any mean estimate after a completed row $n\ge s$ has error greater than $\epsilon$ is at most $Ae^{-c\sqrt s}$, with $A,c$ depending only on $M,\epsilon$.

\subsubsection{Stopping Under a Positive Gap}\label{app:membership-stopping}

Let $g>0$ be a lower bound on the task gap ($\Delta_K$ or $\delta_J$) and set $\epsilon=g/(8\sqrt M)$. Suppose all estimates before a row are accurate to $\epsilon$. For M1/M3, the empirical Top-$K$ set is then $S^\star$, and every false candidate has a queried pair with a violation of at least $g$. For M2/M4, consider the family whose removal certifies the true label. If $J$ is IN, every $K$-set of opponents includes one at least $g$ below $J$. If $J$ is OUT, every $(M-K)$-set includes one at least $g$ above $J$. Thus every hypothesis in this family has a violated pair. Under M4 the sign of the estimated cutoff selects this family on ordinary rows, so its payoffs equal those from the full paired estimates by Appendix~\ref{app:query-preservation}.

In each case let $x$ be the vector being projected, restricted to the query for M3 and taken on the full candidate for M4. Its population counterpart $\theta$ consists of the corresponding means, centered at $\mu_J$ for M4. Since $\|x-\theta\|_\infty\le\epsilon$, the pair violates its empirical constraint by at least $g-2\epsilon$. The distance to this half-space gives $\|r\|_2\ge(g-2\epsilon)/\sqrt2$ for the empirical residual $r$. With $a=2r/\|r\|_1$, Moreau orthogonality gives $r^\top x=\|r\|_2^2$. The conditional score mean given the past and current query uniform is therefore
\[
 a^\top\theta
 \ge \frac{2\|r\|_2^2}{\|r\|_1}-2\epsilon
 \ge \sqrt{2/M}\,(g-2\epsilon)-2\epsilon
 \ge d,\qquad d=\frac{g}{2\sqrt M}.
\]

To apply Lemma~\ref{lem:aq-scalar-crossing}, we define an auxiliary score sequence whose conditional means are positive without conditioning on later events. Fix a false candidate that must be removed and an initial length $s$. Keep its scores through $s$, inserting zeros if deletion or output has already occurred. After $s$, use its actual score when it remains unrejected, no answer has been returned, and the estimates before that row are accurate. Use the constant $d$ otherwise. The resulting scores have conditional means at least $d$ after $s$. On a path with accurate estimates from $s$ onward and with the candidate still live at pending time $t$, its modified account agrees with its actual account through $t$. Such an account has never crossed $h=\log(1/\alpha)$, since processing a crossing either deletes the candidate or ends the procedure.

Take $s=\lfloor\eta t\rfloor$ with $\eta>0$ small enough that $n=t-s$ satisfies the condition in~\eqref{eq:aq-scalar-tail} for all sufficiently large $t$. A pending answer requires at least one of these false candidates to survive. Adding the estimation failure probability to the sum of their crossing tails gives $\Prb(\tau>t)\le Ae^{-c\sqrt t}$. The constants and starting time depend only on $M,K,\alpha,g$. Summation and $\Cost_\tau\le M\tau$ give $\E(\tau+\Cost_\tau)<\infty$.

\subsection{Proof of Proposition~\ref{prop:zero-gap}}\label{app:zero-gap}

Fix a bounded i.i.d.\ score distribution $P$ on $[0,1]^M$ with zero gap for the task. We construct two distributions $Q_1,Q_2$ on $[0,1]^M$ with positive gaps and distinct correct answers $y_1,y_2$, each obtained by independently replacing each row drawn from $P$ with a fixed score vector with probability $\varepsilon\in(0,1)$.

For Top-$K$ identification, $\Delta_K=0$ implies that there exist two distinct $K$-sets $S_1,S_2$ compatible with the mean ordering under $P$. For $j\in\{1,2\}$, use the replacement vector $\mathbf1_{S_j}$ to define $Q_j$. Under $Q_j$, every mean inside $S_j$ exceeds every mean outside $S_j$ by at least $\varepsilon$, so $S_j$ is the unique correct answer. For target qualification, $\delta_J=0$ means that $\mu_J$ equals the $K$-th largest opponent mean under $P$. Define $Q_1$ using the replacement vector $\mathbf1_{\{J\}}$, which gives $J$ score one and every opponent zero, and define $Q_2$ using $\mathbf1-\mathbf1_{\{J\}}$, which reverses those scores. Because multiplying all opponent means by $1-\varepsilon$ and adding a common constant preserves their ordering, the target mean minus the $K$-th largest opponent mean becomes $\varepsilon$ under $Q_1$ and $-\varepsilon$ under $Q_2$, giving positive gap $\varepsilon$ and answers IN and OUT, respectively.

For each $j\in\{1,2\}$, couple the run under $Q_j$ with the run under $P$ using the same original rows, independent replacements across rows, and the same algorithmic randomness. If no replacement occurs in the first $n$ rows, the two runs coincide through row $n$, including all query choices and any output. By a union bound, a replacement occurs within the first $n$ rows with probability at most $n\varepsilon$. Since returning an answer other than $y_j$ is an error under $Q_j$,
\[
 \Prb_P(\tau\le n,\ \widehat y_\tau\ne y_j)
 \le \Prb_{Q_j}(\tau\le n,\ \widehat y_\tau\ne y_j)+n\varepsilon
 \le \alpha+n\varepsilon,\qquad j=1,2.
\]
Letting $\varepsilon\downarrow0$ at fixed $n$ gives $\Prb_P(\tau\le n,\ \widehat y_\tau\ne y_j)\le\alpha$ for $j=1,2$. Because $y_1\ne y_2$, every output differs from at least one of $y_1,y_2$, so summing the two bounds gives $\Prb_P(\tau\le n)\le2\alpha$. Letting $n\to\infty$ yields $\Prb_P(\tau<\infty)\le2\alpha$, and hence $\Prb_P(\tau=\infty)\ge1-2\alpha$.

\subsection{Proof of Theorem~\ref{thm:adaptive-set}}\label{app:adaptive-set}

Let $u$ and $v$ be the models at ranks $K$ and $K+1$. We first derive bounds for a single wealth account. We then apply them to the two candidates that differ only in whether they include $u$ or $v$, and bound the cost of eliminating the remaining candidates.

\subsubsection{Scalar Estimates}\label{app:aq-scalar}

We use the coefficient $b_n$, wealth $W_n$, and curvature constant $\kappa=4/9$ from Lemma~\ref{lem:aq-scalar-crossing}.

We first determine how wealth changes when the mean score is zero or approaches zero. This will identify the time needed to distinguish the difficult pair. The score distribution may vary as its mean changes.

\begin{lemma}\label{lem:aq-scalar-profile}
After at most $s$ arbitrary bounded scores, let the subsequent scores
$Y_1,Y_2,\ldots$ be iid in $[-1,1]$, independently of that prefix.
Fix $s$ and a positive variance lower bound $v$.
If $\E Y_1=0$ and $q=\E Y_1^2\ge v$, then
\begin{equation}\label{eq:aq-scalar-null}
 \frac{\log W_{s+n}}{\log n}\longrightarrow-\frac14
 \quad\text{in probability}.
\end{equation}
The convergence is uniform over these laws and prefixes and over
$n\in[cN,N]$ for any fixed $c>0$ as $N\to\infty$.
If $\E Y_1=d>0$ and $\operatorname{Var}(Y_1)\ge v$, put
$L=\log(1/d)$ and $m=\lfloor d^{-2}L^{-2}\rfloor$.
For fixed $c,C>0$, uniformly over $cm\le n\le Cd^{-2}L$,
\begin{equation}\label{eq:aq-scalar-profile}
 \frac{\log W_{s+n}}L
 =\frac{nd^2}{2qL}-\frac12+o_{\Prb}(1),
 \qquad q=\E Y_1^2.
\end{equation}
\end{lemma}

\begin{proof}
For any complete history $z_1,\ldots,z_n$, concavity gives the deterministic bound
\begin{equation}\label{eq:aq-scalar-leader}
 0\le b_n\le\min\left\{\frac12,
             \frac{(\sum_{j\le n}z_j)_+}{1+\kappa\sum_{j\le n}z_j^2}\right\}.
\end{equation}
For the iid continuation, write $S_n=\sum_{j\le n}Y_j$ and
$V_n=\sum_{j\le n}Y_j^2$. First consider the zero-mean case.
Away from events of exponentially
small probability, $V_n\ge nq/2$ and $|S_n|\le c_vn$.
For large $n$, the maximizing coefficient is then below its upper bound. The identity
$y/(1+by)=y-by^2+b^2y^3/(1+by)$ and the bounded prefix give the
objective derivative
$S_n-bV_n+O(s+1)+O(b^2V_n)$.
The positive-part operation at zero is nonexpansive. Solving this
equation, using \eqref{eq:aq-scalar-leader}, bounds the error relative
to $(S_n)_+/(nq)$ by a constant times
\[
 \frac{s+1}{n}+\frac{(S_n)_+^2}{n^2}
 +\frac{|S_n|\,|V_n-nq|}{n^2}.
\]
The fourth moments of both centered sums are $O(n^2)$ uniformly.
On the exceptional events, the fixed upper bound on the coefficient controls the error.
Consequently
\begin{equation}\label{eq:aq-scalar-root-error}
 \E\left|b_{s+n}-\frac{(S_n)_+}{nq}\right|^2
 \le\frac{C_s}{n^2},\qquad
 \E b_{s+n}^k\le C_s n^{-k/2}\quad(k=2,3,4).
\end{equation}

To sum the quadratic betting cost, put $A_j=(S_j)_+^2/(jq)$.
The bounded-variable central limit estimate is uniform at $q\ge v$;
truncation and the uniform fourth moment imply
$\E A_j\to1/2$ uniformly. For $i\le j$, write $S_j=S_i+T$ with
$T$ independent of $S_i$. The inequality
$|(x+y)_+^2-y_+^2|\le2|x||y|+x^2$ bounds the nonconstant part of
$\E[A_j\mid S_i]$ in $L^2$ by $C\sqrt{i/j}$.
Since $\sup_i\E A_i^2<\infty$, this proves
$|\operatorname{Cov}(A_i,A_j)|\le C\sqrt{i/j}$, and hence
\[
 \operatorname{Var}\left(\sum_{j\le n}\frac{A_j}{j}\right)
 \le C\log(en),\qquad
 \frac1{\log n}\sum_{j\le n}\frac{(S_j)_+^2}{j^2q}
 \longrightarrow\frac12
 \quad\text{in probability}.
\]
By Cauchy--Schwarz, \eqref{eq:aq-scalar-root-error} changes the
quadratic sum by only a bounded $L^1$ error: its $j$-th expected
absolute error is $O_s(j^{-3/2})$.
Thus $q\sum_{j<n}b_{s+j}^2/\log n\to1/2$.
The linear martingale in the exact-log expansion has variance
$O(\log n)$; the centered quadratic martingale has bounded variance;
and the expected absolute cubic remainder is summable by
\eqref{eq:aq-scalar-root-error}. The initial fixed sequence of scores contributes a
bounded amount. This proves \eqref{eq:aq-scalar-null}.
The accumulated quadratic term is nondecreasing. Together with maximal second-moment
inequalities for the martingales, this gives uniform convergence on $[cN,N]$.

For a small positive mean, we determine wealth at time $m$ by comparing with a nearby distribution having mean zero. Construct that distribution as follows. Let
$\psi(a)=\log\E e^{aY_1}$ and $\sigma^2=\psi''(0)\ge v$.
Since $|\psi'''(a)|\le2\psi''(a)$,
$e^{-2|a|}\sigma^2\le\psi''(a)\le e^{2|a|}\sigma^2$.
For sufficiently small $d$, some $0<\theta\le2d/\sigma^2$
satisfies $\psi'(-\theta)=0$.
Tilting by $e^{-\theta Y_1-\psi(-\theta)}$ gives variance at least $v/2$. The relative entropy from the original law to the tilted law is
\[
 \theta d+\psi(-\theta)
 =\int_0^\theta(\theta-a)\psi''(-a)\,da
 \le4d^2/v,
\]
using $\theta\le2d/\sigma^2$ and $e^{2\theta}\le2$ for small $d$.
The product total variation through $m$ observations is therefore
$O(\sqrt{md^2})=o(1)$. Applying \eqref{eq:aq-scalar-null} under this
law and using $\log(cm)/L\to2$ gives
$\log W_{s+\lfloor cm\rfloor}/L\to-1/2$ for every fixed $c>0$.

For later times, let $b_\star$ maximize $\E\log(1+bY_1)$.
Its derivative at zero and curvature imply
$b_\star=d/q+O_v(d^2)$.
Evaluate the empirical derivative at $b_\star$: its centered iid
sum has second moment $O(n)$, and the prefix and regularizer
contribute $O(s+1)$. On $V_n\ge nq/2$, dividing by the objective
curvature of order $n$ proves, also at a constrained endpoint,
\[
 \E(b_{s+n}-d/q)^2
 \le C_v/n+C_v(s+1)^2/n^2+C_vd^4.
\]
The complement again has exponentially small probability.
Summing from $cm$ to $N=Cd^{-2}L$ gives expected squared tracking
loss $O(\log L)=o(L)$.
On this interval $\E\sum b_{s+j}^2=O(L)$, and
\eqref{eq:aq-scalar-leader} with maximal partial-sum bounds gives
$\max_{cm\le j\le N}b_{s+j}=o_{\Prb}(1)$.
Thus the exact-log cubic remainder is $o_{\Prb}(L)$ uniformly in
the endpoint, and the centered linear and quadratic terms have
maximal size $O_{\Prb}(\sqrt L)$.
Finally,
\[
 db-\tfrac12qb^2
 =\frac{d^2}{2q}-\frac q2(b-d/q)^2.
\]
Summing this identity from the starting time used above proves
\eqref{eq:aq-scalar-profile}.
\end{proof}

We next study when either of the two candidates that disagree on the difficult pair is first rejected. The query identities established below show that, after keeping their earlier score histories, their payoffs are selected by the sign of the cumulative observed difference, as specified in the next lemma.

\begin{lemma}\label{lem:aq-scalar-gates}
Let $D\in[-1,1]$ have mean $d>0$, variance at least $v>0$ and second
moment $q$. Fix $\alpha\in(0,1/2)$, set $h=\log(1/\alpha)$,
$L=\log(1/d)$, $H_d=qd^{-2}L$, and
$m=\lfloor d^{-2}L^{-2}\rfloor$.
Let $\rho$ be the first threshold crossing of either of two critical
candidate accounts. Assume the account for the true candidate is a
nonnegative supermartingale. Suppose there is a random time $R$ after which the score representations below apply, with
$\Prb(R>j)\le Ae^{-c\sqrt j}$ uniformly in the law and $d$.
For each deterministic $j$, retain the actual accounts through $j$.
Assume their subsequent scores, on $\{R\le j,\rho\ge t\}$, agree through
row $t$ with the two continuations
\[
 D_n\mathbf1\{w+D_1+\cdots+D_{n-1}>0\},\qquad
 -D_n\mathbf1\{w+D_1+\cdots+D_{n-1}<0\},
 \quad 1\le n\le t-j,
\]
where, conditionally on the fixed-time past, the $D_n$ are iid copies
of $D$, $|w|\le j$, and each retained scalar history has length at
most $j$. The first continuation is the false candidate's account.
Then, uniformly over this class, for each $\epsilon>0$,
\begin{equation}\label{eq:aq-scalar-gate-window}
 \Prb\{m<\rho<(1-\epsilon)H_d\}=o(1),\qquad
 \Prb\{\rho>(1+\epsilon)H_d\}=o(1).
\end{equation}
The variables $\rho/H_d$ are uniformly integrable. In particular,
if $\Prb(\rho\le m)\le2\alpha+o(1)$, then
\begin{equation}\label{eq:aq-scalar-gate-mean}
 1-2\alpha\le\liminf_{d\downarrow0}\frac{\E\rho}{H_d}
 \le\limsup_{d\downarrow0}\frac{\E\rho}{H_d}\le1.
\end{equation}
\end{lemma}

\begin{proof}
Fix $j$ first. After this deterministic row, use the two continuations
without conditioning on $\{R\le j\}$. On that event and while both
accounts survive before the current row, they agree with the actual
accounts, including the row of their first crossing. When an account's sign condition is not met, its payoff is set to zero, so neither its objective nor its wealth changes; omit such rows when counting that account's observations. If its sign condition is met but the observed value is
$D_n=0$, keep that observation. Selection precedes the current score, so each
infinite selected sequence is iid with the corresponding score law,
conditionally on the fixed-time past.

First tilt the future marginal law of $D$ to mean zero, as in the
proof of Lemma~\ref{lem:aq-scalar-profile}. Its product total
variation through physical row $m$ is $O(\sqrt{md^2})=o(1)$.
Under this law the cumulative difference is positive and negative infinitely often, so each account receives infinitely many selected observations. Moreover, for any
$\eta>0$, one can choose $a>0$ so that each account receives at least
$am$ selected observations by row $m$ except with probability $\eta+o(1)$,
uniformly over the laws and bounded offsets $w$.
To justify this lower bound on the number of selected observations, standardize the bounded
increments by their variance. Their maximum size after division
by $\sqrt m$ tends uniformly to zero; their cumulative variance is
deterministic, and fourth-moment bounds give tightness. Thus any
sequence of these laws has the Brownian functional limit.
Brownian paths spend positive time on each side of zero and zero
Lebesgue time at zero, so the two occupation functionals are
continuous at almost every limiting path. A fixed bounded initial
offset vanishes under the same scaling. This proves the assertion.

Apply \eqref{eq:aq-scalar-null} uniformly at the resulting local
indices in $[am,m]$, and transfer back to the original law.
For either account $E^{\pm}$ and every $\eta>0$,
\begin{equation}\label{eq:aq-scalar-entrance}
 \Prb\left\{\rho>m,
       \left|\frac{\log E_m^{\pm}}L+\frac12\right|>\eta\right\}
 \longrightarrow0.
\end{equation}
The proof first fixes $j$. The probability that the stated score representations do not yet apply is at most $Ae^{-c\sqrt j}$. It then lets $j\to\infty$.

Under the original positive-drift law, the cumulative difference is eventually positive by the strong law. The account selected by a positive difference therefore receives infinitely many iid
copies of $D$. For each fixed starting offset, summing concentration
bounds over times gives an expected number $O_j(d^{-2})$ of rows with a nonpositive cumulative difference. Consequently its count of selected observations by row $t$ is
$t+O_{\Prb}(d^{-2})$ for $t$ within bounded multiples of $H_d$.
From the starting time considered above, this count is at least $am$ with arbitrarily high probability.
The wealth estimate \eqref{eq:aq-scalar-profile}, uniform over these observation counts, then
bounds the false account's log wealth divided by $L$ above by
$-\epsilon/2+o_{\Prb}(1)$ throughout
$[m,(1-\epsilon)H_d]$. At $(1+\epsilon)H_d$, on a pending path,
the same profile is $\epsilon/2+o_{\Prb}(1)$.
For the true candidate, apply Ville's inequality conditionally on the observations collected so far:
\[
 \Prb\{\rho>m,\text{ true account crosses after }m\}
 \le\E\left[\mathbf1\{\rho>m\}
                 \min\{1,\alpha E_m^{\rm true}\}\right]=o(1),
\]
where \eqref{eq:aq-scalar-entrance} and boundedness of the integrand
give the last equality. Its selected sequence need not be infinite
under a positive drift. These observations prove
\eqref{eq:aq-scalar-gate-window}.

For integrability, take
$j=\lfloor c_0d^2t\rfloor$ for $t\ge A_0H_d$.
Choose $c_0>0$ sufficiently small and then $A_0$ sufficiently large.
The offset obeys $|w|\le j$ and, for small $d$, is dominated by
the positive mean accumulated before row $t/4$.
On each dyadic interval beginning at $a\ge t/4$, a nonpositive
cumulative difference requires a centered partial sum of size at least
a fixed multiple of $da$. The maximal bounded-increment inequality
and a geometric summation give failure probability $Ce^{-cd^2t}$.
Thus, except with this probability, the account selected by a positive cumulative difference has at least $\lfloor3t/4\rfloor$ iid
selected observations after the initial scores retained above.
Apply~\eqref{eq:aq-scalar-tail} at that deterministic local count.
Here the prefix has length at most $j$, so $dj=O(d^3t)$, which is
absorbed by $d^2t$ for small $d$. The logarithmic term is absorbed
by $t\ge A_0H_d$.
The latter remains true for larger $t$ because $t/\log(et)$ is
increasing. Accounting separately for $\{R>j\}$ gives
\begin{equation}\label{eq:aq-scalar-gate-tail}
 \Prb(\rho>t)\le
 A e^{-c\sqrt{d^2t}}+Ce^{-cd^2t},
 \qquad t\ge A_0H_d.
\end{equation}
Substitute $t=xH_d$ in~\eqref{eq:aq-scalar-gate-tail} and use
$q\ge v$ and $L\ge1$.
The first term is bounded by $Ae^{-c\sqrt{vx}}$, and the exponential term has a uniformly integrable tail as well.
This proves uniform integrability.
Finally \eqref{eq:aq-scalar-gate-window}, the assumed early-crossing
bound and uniform integrability give
\eqref{eq:aq-scalar-gate-mean}.
\end{proof}

\subsubsection{Stopping Time and Query Cost}

Let $H = \{i: \mu_i \ge \mu_u+\gamma\}$ denote the $K-1$ models above $u$, and $L = \{j: \mu_j \le \mu_v-\gamma\}$ the $M-K-1$ models below $v$.
Only two candidates lack a fixed positive violation: $S_u=H\cup\{u\}$
and $S_v=H\cup\{v\}$. Call all others far candidates. Put
$D=X_u-X_v$, $q=\E D^2$ and
$H_{\Delta_K}=q\Delta_K^{-2}\log(1/\Delta_K)$.
For $j\in\{1,3\}$, let $\rho_j$ be the first deletion of either
critical candidate and $F_j$ the earlier of output and removal of
all far candidates. On the first row when a critical account reaches the threshold, its candidate is deleted unless it is left as the sole survivor. In that case the other critical candidate is deleted on the same row, so this first crossing row is $\rho_j$ in either case. We will show that $\E F_j$ stays bounded while
$\E\rho_j$ grows on the scale $H_{\Delta_K}$.
All constants below are uniform at fixed $M,K,\alpha,\gamma,v_0$.

We first bound the time needed to remove the other candidates. For the proof, continue estimators with full queries after output. Let $R_j$ be one
plus the last completed row with any coordinate estimation error
exceeding $\epsilon=\gamma/(8\sqrt M)$. Appendix~\ref{app:sampling-accuracy} gives
\begin{equation}\label{eq:aq-set-calibration}
 \Prb(R_j>s)\le Ae^{-c\sqrt s}.
\end{equation}
Once all coordinate estimates remain accurate, the empirical Top-$K$ set $E$ is $S_u$ or $S_v$.
A far candidate $S$ either omits some $h\in H$, or contains all of
$H$ and one $l\in L$. In the first case pair $h$ with any
$i\in S\setminus E$; in the second pair $l$ with an element of
$E\setminus S$. Both coordinates lie in $E\mathbin\triangle S$, so each
query contains a pair violating $S$ by at least $\gamma$. The projection
bound in Appendix~\ref{app:membership-stopping} with $g=\gamma$ gives
conditional score mean at least $\gamma/(2\sqrt M)$ whenever the estimates are accurate.
Applying its crossing argument to the far candidates and
using~\eqref{eq:aq-set-calibration} gives
\begin{equation}\label{eq:aq-set-cleanup}
 \Prb(F_j>t)\le Ae^{-c\sqrt t},\qquad
 \sup\E(F_j+R_j)<\infty.
\end{equation}

We now consider the two candidates that differ only on the boundary pair. First, we bound their early rejection probability using a nearby distribution under which that pair is tied.
For sufficiently small $d=\E_P D>0$, apply the zero-mean tilt from the proof of Lemma~\ref{lem:aq-scalar-profile} to the whole row. Its density $dQ/dP=e^{-\theta D-\psi(-\theta)}$, where $\psi(a)=\log\E_P e^{aD}$, depends only on $D$. The scalar calculation therefore gives
\begin{equation}\label{eq:aq-tie-law}
 \E_QD=0,\qquad
 \operatorname{KL}(P\Vert Q)\le4d^2/v_0.
\end{equation}
Since $0\le X_i\le1$, Pinsker's inequality gives $\max_i|\E_QX_i-\E_PX_i|\le\sqrt{\operatorname{KL}(P\Vert Q)/2}=O(d)$.
With $d=\Delta_K$, all far gaps under $Q$ remain at least
$\gamma/2$. Both critical cones are true, so Ville's inequality
bounds the probability of either deletion by $2\alpha$.
Pinsker's inequality on the first $m$ raw rows and independent query
uniforms therefore gives
\[
 \Prb(\rho_j\le m)\le2\alpha+\sqrt{2m\Delta_K^2/v_0}
 =2\alpha+o(1),\qquad
 m=\lfloor\Delta_K^{-2}/\log^2(1/\Delta_K)\rfloor.
\]

While both critical candidates survive, M3 queries $u,v$ on every
row because their symmetric difference is contained in the query.
The sign of their empirical difference is the sign of the paired sum, as in M1.
After $R_j$, projection leaves every queried far coordinate unchanged.
The two critical payoffs are exactly
$D_t\mathbf1\{\sum_{s<t}D_s>0\}$ and
$-D_t\mathbf1\{\sum_{s<t}D_s<0\}$, including the first deletion
row. Fix an initial history length and retain both accounts' score histories. The cumulative difference over that history supplies the starting offset for the subsequent sign comparisons.
Equation~\eqref{eq:aq-set-calibration} and
Lemma~\ref{lem:aq-scalar-gates} now yield
\begin{equation}\label{eq:aq-set-boundary}
 1-2\alpha\le\liminf\frac{\E\rho_j}{H_{\Delta_K}}
 \le\limsup\frac{\E\rho_j}{H_{\Delta_K}}\le1.
\end{equation}

These rejection times determine the time and query cost of the actual output. An output requires a critical deletion, and after the far candidates
are gone the first such deletion leaves one survivor. Hence
$0\le\tau_j-\rho_j\le F_j$. By~\eqref{eq:aq-set-cleanup}, the bounds
in~\eqref{eq:aq-set-boundary} also hold for $\tau_j$. Since
$v_0\le q\le1$ and $1-2\alpha>0$, they give the claimed order of
$\E\tau_j$ and the upper bound on $\E\tau_3/\E\tau_1$ in the limit.

Each pending query contains at least two models. After $F_3$ and $R_3$,
an ordinary query contains exactly $u,v$. A full exploration adds at
most $M-2$ queries and occurs with probability $t^{-1/2}$ given
$\{\tau_3\ge t\}$ and the past. Thus
\[
 0\le \E\Cost_{\tau_3}-2\E\tau_3
 \le (M-2)\bigl(\E F_3+\E R_3+2\E\sqrt{\tau_3}\bigr)
 =o(H_{\Delta_K}),
\]
using $\sum_{t\le n}t^{-1/2}\le2\sqrt n$ and Jensen's inequality.
Dividing by $\E\tau_3$ gives $\E\Cost_{\tau_3}/\E\tau_3\to2$.
Since $\Cost_{\tau_1}=M\tau_1$, the query comparison follows.

\subsection{Proof of Theorem~\ref{thm:adaptive-membership}}\label{app:adaptive-membership}

For M2 and M4, the critical comparison is between $J$ and the cutoff
opponent $u$. Write $\widehat\mu_i$ for the average of
$X_{s,i}-X_{s,J}$ over jointly observed rows $s<t$. In M2 these are
the differences of the full coordinate averages.
Let $H = \{i \ne J: \mu_i \ge \max\{\mu_J,\mu_u\}+\gamma\}$ denote the opponents above $u$, and $L = \{i \ne J: \mu_i \le \min\{\mu_J,\mu_u\}-\gamma\}$ the opponents below $u$.
Let $s=\operatorname{sign}(\mu_J-\mu_u)$,
$D=s(X_J-X_u)$ and $q=\E D^2$. Thus $\E D=\delta_J$ and $q\ge v_0$. Use the scale $H_{\delta_J}=q\delta_J^{-2}\log(1/\delta_J)$ from Lemma~\ref{lem:aq-scalar-gates}.
The two critical candidates are $C^+_{H\cup\{u\}}$ and
$C^-_{L\cup\{u\}}$; the first is false for an IN target and the
second for an OUT target. Every other candidate $C_A^+$ contains an
opponent in $L$, and every other candidate $C_B^-$ contains one in
$H$. These are the far candidates.
For $j\in\{2,4\}$, define $\rho_j$ as the first critical deletion
and $F_j$ as the earlier of output and removal of all far candidates.
Each family contains one critical candidate, so emptying a family requires a critical deletion. As in the set proof, the first critical crossing row is therefore $\rho_j$, and $0\le\tau_j-\rho_j\le F_j$.

First, we bound the time needed to remove the far candidates uniformly as the target gap shrinks. For the proof, continue paired estimators with full queries after output. For accuracy
$\gamma/8$, let $R_j$ be one plus the last estimation failure.
Appendix~\ref{app:sampling-accuracy} gives
$\Prb(R_j>a)\le Ae^{-c\sqrt a}$ for both methods, uniformly in $\delta_J$.
For $\delta_J\le\gamma/8$, estimates within the stated error tolerance satisfy
\[
 \widehat\mu_i\ge7\gamma/8\ (i\in H),\qquad
 |\widehat\mu_u|\le\gamma/4,\qquad
 \widehat\mu_i\le-7\gamma/8\ (i\in L).
\]
Thus $u$ is the empirical cutoff opponent.
Set $s_C=1$ for a plus candidate and $s_C=-1$ for a minus candidate.
A queried fixed-gap violator has $-s_C\widehat\mu_i\ge7\gamma/8$.
Thus the pooling root $h=\sum_i(-s_C\widehat\mu_i-h)_+$, summed over
queried opponents in $C$, is at least $7\gamma/16$, so $u$ is inactive.
All positive opponent weights
then belong to fixed-gap violators, and the normalized payoff has
conditional mean at least $\gamma$.
If no such violator is queried, a candidate in the unattempted family
has zero residual, since its signed difference at $u$ is nonpositive.
A far candidate in the attempted family has a violator in its full
support, which the base query includes. Consequently, whenever the estimates are accurate, each far candidate has either zero residual or
conditional payoff mean at least $\gamma$. Full queries activate all
far candidates.

Count only rows with nonzero residual, including those whose realized
payoff is zero. The other rows leave the objective and wealth unchanged.
After deletion or output, append the constant $\gamma$ on every row.
Beyond the first $a$ active scores, replace any score based on
inaccurate estimates by $\gamma$. This gives conditional means at
least $\gamma$ and preserves all live updates on $\{R_j\le a\}$,
since the physical row index is at least the active index.

Between $t/4$ and $t$, both methods query all models at least
$n=\lfloor c_1\sqrt t\rfloor$ times except with probability
$Ae^{-c\sqrt t}$, by Chernoff's inequality for M4.
Take $a=\lfloor c_2 n\rfloor$ with $c_2>0$ small. On $\{R_j\le a\}$,
each far candidate still live before output at time $t$ was active
on all these full queries. By~\eqref{eq:aq-scalar-tail}, the probability
that its modified account has yet to cross at active index $n$ is at most $e^{-cn}$.
A union bound over far candidates, adding $\Prb(R_j>a)$, gives
\begin{equation}\label{eq:aq-member-cleanup}
 \Prb(F_j>t)\le Ae^{-ct^{1/4}},\qquad
 \sup\E(F_j+R_j)<\infty.
\end{equation}

We next analyze the target and its cutoff opponent. Once the estimates remain accurate and before the first critical deletion, the attempted
critical candidate has support $\{u\}$ whenever the cutoff is
nonzero. At zero cutoff the full-query fallback also observes $u$.
Thus both M2 and M4 query $J,u$ and have critical payoffs
\[
 (X_{t,J}-X_{t,u})\mathbf1\{\widehat\mu_u<0\},\qquad
 -(X_{t,J}-X_{t,u})\mathbf1\{\widehat\mu_u>0\}.
\]
This includes the deletion row. M4 may omit $u$ earlier:
for a deterministic row $a$, retain both scalar histories and the
signed observed pair sum $w$, where $|w|\le a$.
Future raw differences are iid conditional on this fixed-time past.
The continued score sequences selected by the sign of $w+\sum_{k=a+1}^{t-1}D_k$ agree with
the actual accounts on $\{R_j\le a,\rho_j\ge t\}$.
Thus the actual accounts satisfy the representation required by Lemma~\ref{lem:aq-scalar-gates} after retaining their initial histories.

Apply the whole-vector tilt in~\eqref{eq:aq-tie-law} to
$s(X_J-X_u)$. Under the resulting tied law both critical cones are
true and the far gaps remain positive. Exactly as in the set proof,
Ville and Pinsker give $\Prb(\rho_j\le m)\le2\alpha+o(1)$ for
$m=\lfloor\delta_J^{-2}/\log^2(1/\delta_J)\rfloor$.
Lemma~\ref{lem:aq-scalar-gates}, \eqref{eq:aq-member-cleanup} and
$0\le\tau_j-\rho_j\le F_j$ imply
\begin{equation}\label{eq:aq-member-boundary}
 1-2\alpha\le
 \liminf\frac{\E\tau_j}{H_{\delta_J}}
 \le\limsup\frac{\E\tau_j}{H_{\delta_J}}\le1.
\end{equation}

Since $v_0\le q\le1$ and $1-2\alpha>0$, these bounds give the stated
order of $\E\tau_j$. Taking the upper bound for $j=4$ and the lower
bound for $j=2$ gives the stopping-time comparison.

Finally, we count the additional queries caused by refreshing uncertain estimates, querying all models at a zero cutoff, and random exploration. After $N\ge2$ observations, bounded paired differences give
$2\widehat{\mathrm{SE}}_i\le2/\sqrt{N-1}$. Let $G$ be the maximum of
$F_4$, $R_4$ and the row of the $(2+\lceil16/\gamma^2\rceil)$-th full
exploration, continuing the independent exploration draws after output.
By Chernoff's inequality for this fixed exploration count
and~\eqref{eq:aq-member-cleanup}, $\sup_{\delta_J}\E G<\infty$.
On each pending row after $G$, only the critical candidates remain
and every refresh margin is at most $\gamma/2$. At a nonzero cutoff,
the attempted candidate has opponent support $\{u\}$. Every other
opponent $i$ in that candidate satisfies
$-s_C\widehat\mu_i+2\widehat{\mathrm{SE}}_i
\le-7\gamma/8+\gamma/2<0<-s_C\widehat\mu_u$.
Hence an ordinary query contains exactly $J,u$ unless the cutoff is zero.

To count zero cutoffs, write $S_n$ for the sum of the first $n$ observed
values of $D$, continuing with full queries after output. These observations
are iid by predictable sampling. Each pending row after $G$ observes $u$,
so each fallback uses a distinct local index $n$ with $S_n=0$.
By Hoeffding's inequality their expected number is bounded by
$1+\sum_{n\ge1}e^{-n\delta_J^2/2}\le1+2\delta_J^{-2}$.
As in M3, the expected number of explorations through $\tau_4$ is at
most $2\E\sqrt{\tau_4}$. Every pending query contains $J$ and an
opponent, since a nonzero cutoff gives a nonzero opponent residual in
the selected family and a zero cutoff triggers a full query. Therefore
\[
 0\le\E\Cost_{\tau_4}-2\E\tau_4
 \le(M-2)\bigl(\E G+1+2\delta_J^{-2}+2\E\sqrt{\tau_4}\bigr)
 =o(H_{\delta_J}),
\]
by~\eqref{eq:aq-member-boundary} and Jensen's inequality. Dividing by
$\E\tau_4$ gives $\E\Cost_{\tau_4}/\E\tau_4\to2$, and
$\Cost_{\tau_2}=M\tau_2$ yields the query comparison.

\subsection{Proof of Theorem~\ref{thm:separation}}\label{app:membership-separation}

We first bound the cost of direct target qualification. For $j\in\{2,4\}$ and $\delta_J\ge\delta>0$, the constants in the stopping bound of Appendix~\ref{app:membership-stopping} depend only on $M,K,\alpha,\delta$. Summing that bound and using $C_j\le M\tau_j$ gives the asserted uniform bound on $\E C_j$.

We next derive a lower bound for Top-$K$ identification. Fix a row law $P$ with true set $S^\star$ and boundary gap $\Delta_K>0$. We construct a law with a different Top-$K$ set and bound the information each query provides for distinguishing the two laws.

Choose $i\in S^\star$ and $j\notin S^\star$ at the boundary, so $\mu_i-\mu_j=\Delta_K$, and put $S'=(S^\star\setminus\{i\})\cup\{j\}$. For $\Delta_K/(1+\Delta_K)<\eta<1$, let $Q_\eta=(1-\eta)P+\eta\delta_{\mathbf1_{S'}}$, where $\mathbf1_{S'}$ is the indicator vector of $S'$. Under $Q_\eta$, the smallest mean on $S'$ exceeds the largest mean outside it by $\eta-(1-\eta)\Delta_K>0$. Thus $S'$ is its unique Top-$K$ set.

For each queried set $A$, the marginal $Q_{\eta,A}$ contains $P_A$ with weight $1-\eta$, so $\operatorname{KL}(P_A\Vert Q_{\eta,A})\le-\log(1-\eta)\mathbf1\{A\ne\varnothing\}$. Let $A_n$ be the event that the comparator returns $S^\star$ by row $n$. Query and stopping decisions use the same conditional kernels under both laws. The chain rule through $\tau\wedge n$ and data processing for $A_n$ therefore give~\cite{kaufmann2016complexity}
\[
 \operatorname{kl}\{P(A_n),Q_\eta(A_n)\}
 \le -\log(1-\eta)\,\E_P\sum_{t\le\tau\wedge n}\mathbf1\{Q_t\ne\varnothing\}.
\]
Almost sure return and error control give $\lim_n P(A_n)\ge1-\alpha$ and $\lim_n Q_\eta(A_n)\le\alpha$. Pass to the limit in $n$ by lower semicontinuity on the left and monotone convergence on the right. For $p\ge1-\alpha>\alpha\ge q$, $\operatorname{kl}(p,q)\ge\operatorname{kl}(1-\alpha,\alpha)$ by monotonicity in each argument. Now let $\eta\downarrow\Delta_K/(1+\Delta_K)$. This gives $\E_P\sum_{t\le\tau}\mathbf1\{Q_t\ne\varnothing\}\ge\operatorname{kl}(1-\alpha,\alpha)/\log(1+\Delta_K)$. Each such row costs at least one model query and costs $M$ under full observation. Since $\log(1+\Delta_K)\le\Delta_K$, both claimed bounds follow.

\section{Polynomial FTRL Implementation}\label{app:ftrl-implementation}

To compute the coefficient in constant time per update as the score history grows, replace $\log(1+u)$ in~\eqref{eq:ftrl-opt} by $P_{32}(u)=\sum_{k=1}^{32}(-1)^{k+1}u^k/k$. The 32 sums $\sum_{j\le n}z_j^k$ determine the regularized objective and its maximizer
\[
 F_n(b)=\sum_{j\le n}P_{32}(bz_j)-R(b),\qquad
 \tilde b_n=\arg\max_{0\le b\le1/2}F_n(b).
\]
The identity $P_{32}'(u)=(1-u^{32})/(1+u)$ gives $|P_{32}'|\le2$ and $-P_{32}''\ge2/5$ on $[-1/2,1/2]$. Endpoint checks and bisection thus determine the unique maximizer to any fixed precision in constant time. The bounds below use the maximizer $\tilde b_n$. Under a true hypothesis, $\E[z_t\mid\mathcal F_{t-1}]\le0$. Since $\tilde b_{t-1}\in[0,1/2]$ is determined by past scores, the factor $1+\tilde b_{t-1}z_t$ is nonnegative and has conditional mean at most one. Thus the actual wealth $W_n=\prod_{j\le n}(1+\tilde b_{j-1}z_j)$ satisfies the same Ville bound as in Theorem~\ref{thm:valid}.

\subsection{Crossing Times}
We show that the tail bound~\eqref{eq:aq-scalar-tail} holds for $W_n$. Put $V_n=\sum_{j\le n}z_j^2$ and $\Phi_n=F_n(\tilde b_n)$. The argument bounding changes in the coefficient in Lemma~\ref{lem:aq-scalar-crossing}, with curvature $2/5$, shows that the cumulative polynomial return is within $16\log(1+V_n)$ of the regularized maximum. Since $|P_{32}(u)-\log(1+u)|\le2|u|^{33}/33$,
\[
 \Phi_n-\log W_n\le16\log(1+V_n)
   +\frac2{33}\sum_{j\le n}\tilde b_{j-1}^{33}z_j^2.
\]
Before the first crossing of $h$, it suffices to bound $\Phi_n$ by $O_h(1+\log(1+V_n))$. To control the cumulative Taylor error, put $H=1+\max_{j\le n}\Phi_j$. Strong concavity and $F_j(0)=0$ imply $\Phi_j\ge V_j\tilde b_j^2/5$. Thus $\tilde b_j^{33}\le2^{-29}f(V_j)$, where $f(v)=\min\{1,(5H/v)^2\}$ and $f(0)=1$. Since $f$ decreases and $V_j-V_{j-1}=z_j^2\le1$, its left sum exceeds its integral by at most $f(0)=1$. Hence the error sum, including the factor $2/33$, is at most $(2/33)2^{-29}(1+\int_0^\infty f(v)\,dv)=(2/33)2^{-29}(1+10H)\le H/100$. On a path still below the threshold through row $n$, taking the maximum in the display gives $H\le1+h+16\log(1+V_n)+H/100$.

Consider the fixed coefficient $\beta=d/2$. Since $P_{32}(u)\ge u-u^2$ on $[-1/2,1/2]$, the comparator loses at most $ds$ during a prefix of length $s$. Each subsequent score with conditional mean at least $d$ gives conditional expected polynomial return at least $d^2/4$. Its centered returns are bounded in absolute value by $2d$. Applying the Azuma argument of Lemma~\ref{lem:aq-scalar-crossing} with the preceding bound on $\Phi_n$ proves~\eqref{eq:aq-scalar-tail}, with a possibly larger $A_h$. This also proves the stopping and uniform cost bounds for the polynomial accounts.

\subsection{Query Costs at Small Gaps}
To show that the polynomial update has the same limiting wealth behavior as in Lemma~\ref{lem:aq-scalar-profile}, compare $\tilde b_n$ with the exact maximizer $b_n$ on the same history. At $\tilde b_n$, the exact and polynomial objective derivatives differ by at most $2\tilde b_n^{32}V_n$. Using curvature at least $1+4V_n/9$ for the exact objective and $1+2V_n/5$ for the polynomial objective, constrained optimality and $F_n'(0)=\sum_{j\le n}z_j$ give
\[
 |b_n-\tilde b_n|\le\frac{2\tilde b_n^{32}V_n}{1+4V_n/9}
 \le\tfrac92\tilde b_n^{32},\qquad
 \tilde b_n\le\frac{(\sum_{j\le n}z_j)_+}{1+2V_n/5}.
\]
After a prefix of length at most $s$, let the subsequent iid scores have mean $d\ge0$ and second moment at least $v>0$. Their centered sum through $j$ observations has fourth moment $O(j^2)$, and their sum of squares is at least $jv/2$ except with exponentially small probability. The second bound therefore gives $\E\tilde b_{s+j}^4\le C_{s,v}(j^{-2}+d^4)$ for a prefix of length $s$, uniformly also for shorter prefixes. Writing $W_n^{\log}$ for the exact account on the same history, the first bound and $|\partial_b\log(1+bz)|\le2$ give
\[
 \E\max_{n\le s+N}|\log W_n-\log W_n^{\log}|
 \le9\,\E\sum_{j<s+N}\tilde b_j^{32}\le C_{s,v}(1+Nd^4).
\]
The right-hand side stays bounded when $d=0$ or $N=O(d^{-2}\log(1/d))$ as $d\downarrow0$. After division by the growing normalizing factors in~\eqref{eq:aq-scalar-null} and~\eqref{eq:aq-scalar-profile}, this bounded difference vanishes. Thus the same limits hold for $W_n$. Zero scores leave both accounts unchanged. Lemma~\ref{lem:aq-scalar-gates} now applies with these profiles and~\eqref{eq:aq-scalar-tail}, yielding the query comparisons in Theorems~\ref{thm:adaptive-set} and~\ref{thm:adaptive-membership}.

\section{Experimental Details and Additional Results}\label{app:additional-controls}

\subsection{Implementation and Reporting}
All PACE methods compute FTRL betting coefficients via the degree-32 Taylor polynomial $P_{32}$ from Appendix~\ref{app:ftrl-implementation}. For PrPl-EB~\cite{10.1093/jrsssb/qkad009}, we apply the confidence sequence to the rescaled score difference $(X_i-X_j+1)/2$ on each question. We divide $\alpha$ equally among all model pairs for Top-$K$ identification and among the $M-1$ pairs involving $J$ for target qualification.

For each ordered pair of models, RankCS averages the five wealth processes obtained by betting fixed fractions $0.03$, $0.06$, $0.12$, $0.25$, and $0.5$, with equal weight on each process. With retirement, a pair's wealth stays at its last value once either model retires. For larger pools we use the equivalent integer-program formulation of the weak-order test~\cite{khosravi2026rank}. Single-arm bandit baselines draw one new question per model pull, whereas BRR and VRR query all active models on a shared question. In the synthetic experiments of Section~\ref{sec:experiments}, model columns are randomly permuted once per stream. In the Open LLM Leaderboard benchmarks from tinyBenchmarks~\cite{polo2024tinybenchmarks}, item counts are 14,042 for MMLU, 10,042 for HellaSwag, 1,319 for GSM8K, 817 for TruthfulQA, 1,267 for WinoGrande, and 1,172 for ARC-Challenge. Mean total queries $\Cost_\tau$ and mean questions $\tau$ include runs whose first output is wrong. Empirical error rates use all scheduled runs as the denominator. If a baseline fails to return on any stream within budget, we omit its mean cost. The leaderboard tables and figures report the number of returns for those baselines.

\subsection{Open LLM Leaderboard Results}
Table~\ref{tab:app-e6-b15-full} gives the mean query counts for Top-5 identification on the six leaderboard benchmarks, including RankCS and RankCS with retirement.

\begin{table}[ht]
\centering
\small
\setlength{\tabcolsep}{2.6pt}
\caption{Mean total model queries for Top-5 identification in the 15-model pools, rounded to the nearest query. Each benchmark uses 100 paired streams with a budget of 100,000 questions and $\alpha=0.05$. When a method returned on fewer than 100 streams, the table gives the return count in parentheses. All four PACE methods returned on all 100 streams with zero errors on every benchmark and task.}
\label{tab:app-e6-b15-full}
\vspace{4pt}
\begin{tabular}{lrrrrrrrr}\toprule
Benchmark & $\Delta_5$ & M3 & M1 & RankCS & \shortstack{RankCS\\with retirement} & PrPl-EB & BRR & VRR\\\midrule
MMLU & 0.0097 & \textbf{52,630} & 286,979 & 489,026 & 83,441 & -- (68/100) & -- (0/100) & -- (0/100)\\
HellaSwag & 0.0073 & \textbf{35,201} & 134,734 & 234,728 & 56,569 & 428,349 & -- (0/100) & -- (0/100)\\
GSM8K & 0.1039 & \textbf{1,576} & 4,956 & 9,272 & 2,451 & 14,461 & 19,019 & 14,342\\
TruthfulQA & 0.0163 & \textbf{14,214} & 65,624 & 115,372 & 22,894 & 223,026 & -- (71/100) & -- (0/100)\\
WinoGrande & 0.0134 & \textbf{36,652} & 116,128 & 201,738 & 56,821 & 426,274 & -- (3/100) & -- (0/100)\\
ARC-Challenge & 0.0128 & \textbf{27,351} & 127,419 & 226,886 & 44,384 & -- (99/100) & -- (1/100) & -- (0/100)\\
\bottomrule\end{tabular}
\end{table}

\begin{figure}[htbp]
\centering
\includegraphics[width=\linewidth]{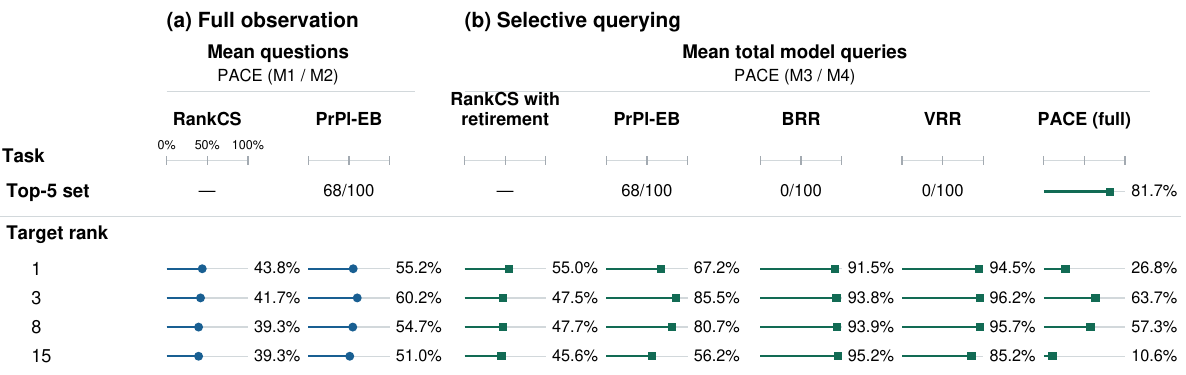}
\caption{Percentage reductions in mean questions and mean total model queries on MMLU, as in Figure~\ref{fig:e6-b15}. Dashes refer to the Top-5 results against RankCS in Figure~\ref{fig:leaderboard-query-savings}. Return counts replace percentages when a comparator returned on fewer than 100 streams.}
\label{fig:app-leaderboard-costs}
\end{figure}

\begin{figure}[htbp]
\centering
\includegraphics[width=\linewidth]{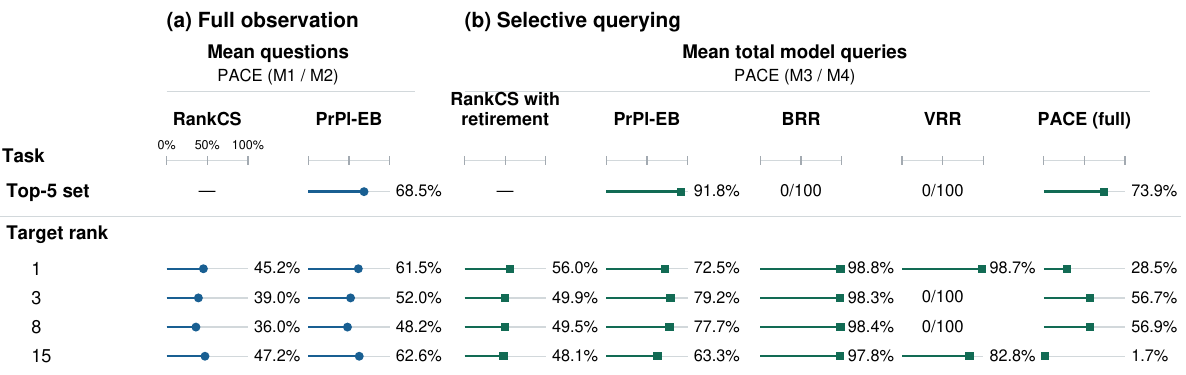}
\caption{Percentage reductions in mean questions and mean total model queries on HellaSwag, as in Figure~\ref{fig:e6-b15}. Dashes refer to the Top-5 results against RankCS in Figure~\ref{fig:leaderboard-query-savings}. Return counts replace percentages when a comparator returned on fewer than 100 streams.}
\label{fig:app-savings-hellaswag}
\end{figure}

\begin{figure}[htbp]
\centering
\includegraphics[width=\linewidth]{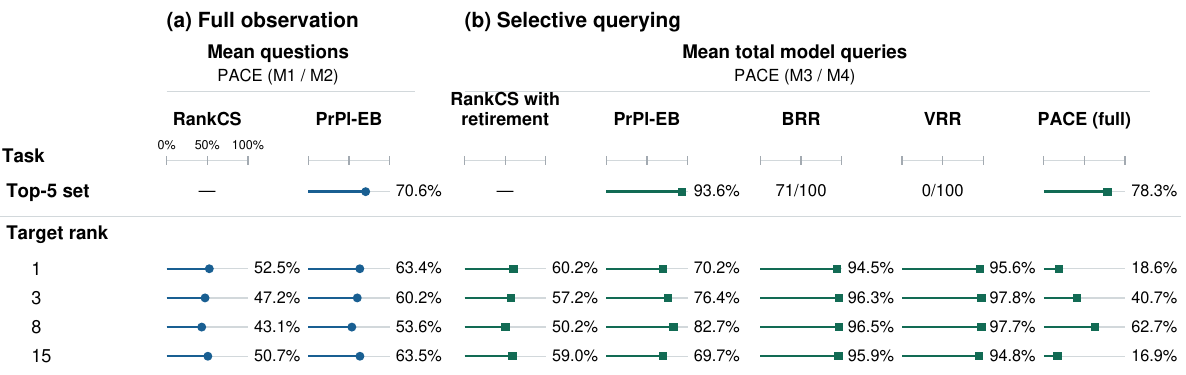}
\caption{Percentage reductions in mean questions and mean total model queries on TruthfulQA, as in Figure~\ref{fig:e6-b15}. Dashes refer to the Top-5 results against RankCS in Figure~\ref{fig:leaderboard-query-savings}. Return counts replace percentages when a comparator returned on fewer than 100 streams.}
\label{fig:app-savings-truthfulqa}
\end{figure}

\begin{figure}[htbp]
\centering
\includegraphics[width=\linewidth]{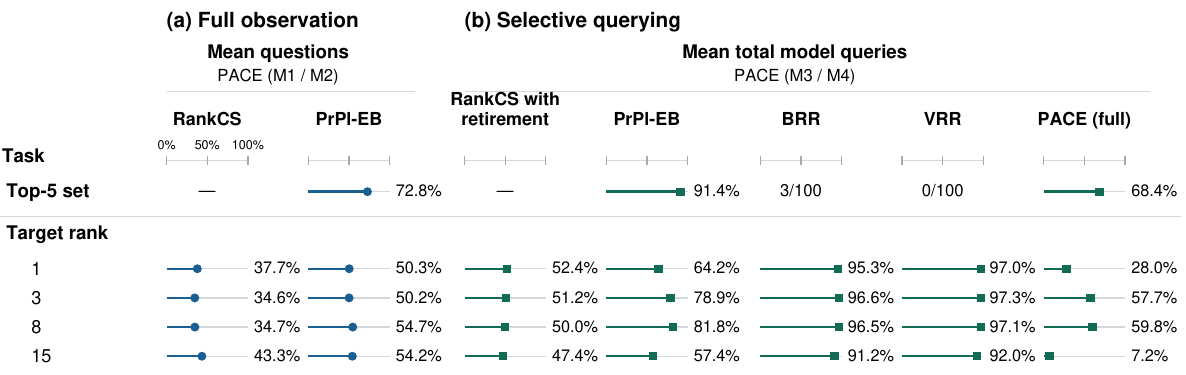}
\caption{Percentage reductions in mean questions and mean total model queries on WinoGrande, as in Figure~\ref{fig:e6-b15}. Dashes refer to the Top-5 results against RankCS in Figure~\ref{fig:leaderboard-query-savings}. Return counts replace percentages when a comparator returned on fewer than 100 streams.}
\label{fig:app-savings-winogrande}
\end{figure}

\begin{figure}[htbp]
\centering
\includegraphics[width=\linewidth]{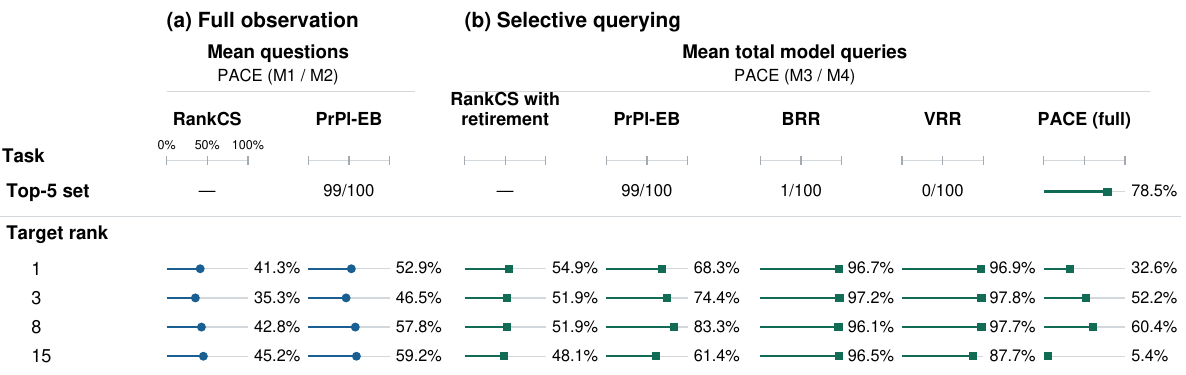}
\caption{Percentage reductions in mean questions and mean total model queries on ARC-Challenge, as in Figure~\ref{fig:e6-b15}. Dashes refer to the Top-5 results against RankCS in Figure~\ref{fig:leaderboard-query-savings}. Return counts replace percentages when a comparator returned on fewer than 100 streams.}
\label{fig:app-savings-arc-challenge}
\end{figure}

\subsection{Negative Correlation Across the Cutoff}
We use the 12-model pool from Section~\ref{sec:exp-e2} with $K=3$, $\Delta_K=0.05$, and $\rho=0.7$. We generate scores as
\[
 X_{t,i}=\mathbf{1}\!\left\{s_i\sqrt{\rho}\,Z_t+\sqrt{1-\rho}\,\varepsilon_{t,i}\leq\Phi^{-1}(\mu_i)\right\},
\]
where $s_i=+1$ for the top $K$ models and $s_i=-1$ for the others.

Under this negative dependence, full-observation PACE uses $36.7\%$ fewer questions than RankCS for Top-3 identification and $34.9\%$ fewer for qualification of the rank-3 target. For the same tasks, selective PACE uses $29.2\%$ and $25.2\%$ fewer model queries, respectively, than RankCS with retirement (Table~\ref{tab:opposed-costs}).

\begin{table}[htbp]
\centering
\small
\caption{Mean costs to the first answer under negative dependence across the cutoff, with $M=12$, $K=3$, $\Delta_K=0.05$, and $\rho=0.7$. All methods use the same 100 streams.}
\label{tab:opposed-costs}
\footnotesize
\setlength{\tabcolsep}{2.8pt}
\begin{tabular}{@{}lrrr@{\hspace{6pt}}rrrrrrr@{}}\toprule
 & \multicolumn{3}{c}{Mean questions} & \multicolumn{7}{c}{Mean total model queries}\\
\cmidrule(lr){2-4}\cmidrule(lr){5-11}
Task & \shortstack{PACE\\(full)} & RankCS & PrPl-EB & \shortstack{PACE\\(full)} & \shortstack{PACE\\(selective)} & \shortstack{RankCS\\(retirement)} & KL-LUCB & LUCB++ & BRR & VRR\\\midrule
Top-3 set & 3,394 & 5,363 & 13,879 & 40,724 & 8,358 & 11,805 & 42,966 & 22,800 & 39,534 & 46,843\\
Rank-3 target & 3,489 & 5,361 & 11,116 & 41,870 & 8,818 & 11,781 & & & 39,534 & 46,843\\\bottomrule\end{tabular}
\end{table}

\subsection{Effect of the RankCS Betting Rule}\label{app:rankcs-ftrl}

We repeat every RankCS comparison with the FTRL update used by PACE (Appendix~\ref{app:ftrl-implementation}). Each ordered pair in RankCS chooses its betting fraction from its own past score differences. Both versions of RankCS use the same streams, budgets, weak-order tests, and stopping rules. Under selective querying, both apply the same retirement rule.

Tables~\ref{tab:rankcs-ftrl-benchmarks} and~\ref{tab:rankcs-ftrl-synthetic} give the reduction in mean cost for PACE at each setting. With FTRL for both methods, PACE uses fewer questions under full observation and fewer model queries under selective querying in every comparison in the two tables. On the six 15-model leaderboard pools, M1 saves $46.3\%$ to $49.7\%$ of questions for Top-5 identification, and M3 saves $33.2\%$ to $43.3\%$ of model queries. Figure~\ref{fig:e6-extended} shows how changing the RankCS betting rule affects returns on the two near-tied pools.

\begin{table}[H]
\centering
\footnotesize
\setlength{\tabcolsep}{4.5pt}
\caption{Reduction in mean cost for PACE relative to RankCS when both use FTRL on the six 15-model leaderboard pools. Each entry compares 100 paired streams. The numbered columns give the target rank for qualification.}
\label{tab:rankcs-ftrl-benchmarks}
\begin{tabular}{@{}lrrrrr@{\hspace{1.2em}}rrrrr@{}}\toprule
 & \multicolumn{5}{c}{Full observation (questions)} & \multicolumn{5}{c}{Selective querying (model queries)}\\
\cmidrule(lr){2-6}\cmidrule(lr){7-11}
Benchmark & Top-5 & 1 & 3 & 8 & 15 & Top-5 & 1 & 3 & 8 & 15\\\midrule
MMLU & 47.5\% & 39.9\% & 43.6\% & 41.6\% & 33.7\% & 42.4\% & 51.6\% & 49.0\% & 49.2\% & 40.5\%\\
HellaSwag & 46.3\% & 37.7\% & 38.9\% & 34.0\% & 38.0\% & 40.2\% & 50.0\% & 48.5\% & 47.2\% & 39.1\%\\
GSM8K & 46.9\% & 45.5\% & 41.5\% & 44.3\% & 46.6\% & 33.2\% & 57.6\% & 52.8\% & 60.3\% & 63.6\%\\
TruthfulQA & 47.8\% & 45.3\% & 42.3\% & 43.5\% & 41.8\% & 40.5\% & 54.2\% & 52.4\% & 48.6\% & 51.6\%\\
WinoGrande & 48.0\% & 35.7\% & 38.8\% & 38.6\% & 37.8\% & 39.6\% & 50.7\% & 53.5\% & 52.0\% & 42.3\%\\
ARC-Challenge & 49.7\% & 37.1\% & 34.0\% & 44.0\% & 35.2\% & 43.3\% & 51.4\% & 49.9\% & 52.0\% & 38.7\%\\
\bottomrule\end{tabular}
\end{table}

\begin{table}[H]
\centering
\small
\caption{Reduction in mean cost for PACE relative to RankCS when both use FTRL on the synthetic experiments. Each entry compares 100 paired streams at the indicated setting. Full compares M1 or M2 with RankCS by questions. Selective compares M3 or M4 with RankCS using retirement by total model queries.}
\label{tab:rankcs-ftrl-synthetic}
\begin{tabular}{@{}lr rr rr@{}}\toprule
 & & \multicolumn{2}{c}{Top-3 set} & \multicolumn{2}{c}{Rank-3 target}\\
\cmidrule(lr){3-4}\cmidrule(lr){5-6}
Experiment & Value & Full & Selective & Full & Selective\\\midrule
Correlation ($\rho$) & 0 & 46.5\% & 37.3\% & 45.9\% & 36.1\%\\
 & 0.2 & 51.0\% & 41.3\% & 51.7\% & 41.6\%\\
 & 0.4 & 53.6\% & 43.7\% & 52.2\% & 41.6\%\\
 & 0.6 & 53.5\% & 43.1\% & 52.2\% & 40.2\%\\
 & 0.8 & 44.4\% & 33.5\% & 44.4\% & 32.1\%\\
 & 0.9 & 50.4\% & 39.0\% & 50.6\% & 38.2\%\\
Opposite signs across cutoff ($\rho$) & 0.7 & 45.3\% & 37.5\% & 43.7\% & 34.0\%\\
\addlinespace
Boundary gap ($\Delta_K$) & 0.12 & 46.0\% & 35.5\% & 47.0\% & 37.4\%\\
 & 0.08 & 53.1\% & 40.2\% & 54.1\% & 41.6\%\\
 & 0.05 & 47.1\% & 37.8\% & 47.6\% & 37.5\%\\
 & 0.03 & 49.3\% & 42.8\% & 48.3\% & 40.2\%\\
 & 0.02 & 44.1\% & 39.4\% & 43.8\% & 37.9\%\\
 & 0.015 & 46.6\% & 43.2\% & 45.8\% & 41.4\%\\
\addlinespace
Pool size ($M$) & 6 & 38.6\% & 32.7\% & 37.2\% & 30.9\%\\
 & 12 & 49.0\% & 37.9\% & 48.8\% & 36.9\%\\
 & 20 & 56.3\% & 40.9\% & 56.4\% & 39.3\%\\
 & 40 & 63.0\% & 40.0\% & 63.1\% & 39.3\%\\
 & 60 & 65.9\% & 40.0\% & 66.1\% & 38.0\%\\
\bottomrule\end{tabular}

\smallskip
\begin{tabular}{@{}l rr rr rr@{}}\toprule
 & \multicolumn{2}{c}{Top-3 set} & \multicolumn{2}{c}{Rank-1 target} & \multicolumn{2}{c}{Rank-10 target}\\
\cmidrule(lr){2-3}\cmidrule(lr){4-5}\cmidrule(lr){6-7}
Task separation ($\Delta_K$) & Full & Selective & Full & Selective & Full & Selective\\\midrule
0.1 & 46.4\% & 35.4\% & 43.8\% & 54.8\% & 41.2\% & 59.1\%\\
0.05 & 52.3\% & 41.6\% & 42.7\% & 55.1\% & 42.7\% & 59.8\%\\
0.02 & 53.0\% & 47.7\% & 47.5\% & 58.5\% & 41.0\% & 59.0\%\\
0.01 & 51.3\% & 48.5\% & 44.4\% & 55.3\% & 42.2\% & 59.1\%\\
0.005 & 55.6\% & 54.2\% & 45.6\% & 57.2\% & 41.5\% & 59.8\%\\
\bottomrule\end{tabular}
\end{table}

\section{Empirical First-Output Errors}\label{app:error-experiments}

\begin{table}[htbp]
\centering
\small
\caption{Empirical first-output error rates for PACE at $\alpha=0.05$. Each fraction counts incorrect Top-$K$ sets or IN/OUT decisions among 100 scheduled streams, separately for each method and setting. All PACE runs returned an answer.}
\label{tab:pace-first-output-errors}
\begin{tabular}{@{}llr@{}}\toprule
Experiment & Setting & Error rate\\\midrule
Score correlation & $\rho=0.2$ & $1/100$\\
 & All other settings, including negative dependence & $0/100$\\
\addlinespace
Boundary gap and pool size & $M=10$, $\Delta_K=0.03$ & $2/100$\\
 & All other settings & $0/100$\\
\addlinespace
Task separation & All gaps and tasks & $0/100$\\
Leaderboard, 15-model pools & All six benchmarks and all tasks & $0/100$\\
\addlinespace
Near-tied 20-model pools (M1) & MMLU & $1/100$\\
 & HellaSwag & $0/100$\\\bottomrule
\end{tabular}
\end{table}

\section{Testing General Order Constraints}\label{app:order-constraints}

To apply PACE to different ranking tasks under full observation, express each candidate answer through the comparisons it requires. For example, the claim that a model ranks first in a given group compares it with every other member but leaves the order among those members unrestricted. A proposed complete ranking $i_1,\ldots,i_M$ specifies the chain $\theta_{i_1}\ge\cdots\ge\theta_{i_M}$. In either case, let $C$ be the intersection of the half-spaces $\{\theta:\theta_i\ge\theta_j\}$ for the specified comparisons. The candidate hypothesis is $\mu\in C$.

To test this hypothesis, project the empirical means from past questions onto $C$. The projection is the closest mean vector consistent with the proposed order. Its residual supplies the weights for the next question in~\eqref{eq:projection-payoff}. For example, if the proposed order is $1,2,3$ from highest to lowest and the empirical means are $(0.8,0.4,0.6)$, the projection is $(0.8,0.5,0.5)$ and the next payoff is $X_{t,3}-X_{t,2}$. Since the order cone is closed, convex and contains all constant vectors, Appendix~\ref{app:validity} gives a payoff in $[-1,1]$ with nonpositive conditional mean under the hypothesis. The probability that its wealth ever reaches $1/\alpha$ is therefore at most $\alpha$.

To recover an unknown complete ranking, use each of the $M!$ permutations as a candidate and maintain one wealth account for its order cone. Apply the deletion rule of Section~\ref{sec:set-output} and report the last remaining permutation. If the population means are all distinct, exactly one candidate is true. Reporting a wrong ranking requires deleting that candidate, so the error probability is at most $\alpha$. In every false permutation, some model appears immediately ahead of a model with a higher population mean. Their means differ by at least the smallest gap between consecutive models in the true ranking. The crossing argument in Appendix~\ref{app:membership-stopping} thus applies to each false permutation. Since there are finitely many and at least one survives while the procedure is pending, the expected number of questions until a ranking is reported is finite.

\end{document}